\documentclass[10pt,twocolumn,letterpaper]{article}
\usepackage{cvpr}
\definecolor{cvprblue}{rgb}{0.21,0.49,0.74}
\usepackage[breaklinks,colorlinks,allcolors=cvprblue]{hyperref}
\usepackage[utf8]{inputenc}
\usepackage[T1]{fontenc}
\usepackage{url}
\usepackage{booktabs}
\usepackage{amsfonts}
\usepackage{nicefrac}
\usepackage{microtype}
\usepackage{amsmath} 
\usepackage{graphicx}
\usepackage{array}
\usepackage{booktabs}
\usepackage[table,xcdraw]{xcolor}
\usepackage[normalem]{ulem}
\useunder{\uline}{\ul}{}
\usepackage{comment}
\usepackage{wasysym}
\usepackage{multirow}
\usepackage{subcaption}
\usepackage{amssymb}
\usepackage{makecell}
\usepackage{pifont}
\newcommand{\cmk}{\ding{51}}
\newcommand{\xmk}{\ding{55}}

\newcommand{\semcolor}[1]{\textcolor{#1}{\rule{1.2ex}{1.2ex}}}
\definecolor{road}{rgb}{0.502,0.000,0.502}
\definecolor{walkway}{rgb}{0.800,0.639,0.282}
\definecolor{dirt}{rgb}{0.502,0.000,0.000}
\definecolor{gravel}{rgb}{0.753,0.753,0.753}
\definecolor{rock}{rgb}{0.965,0.471,0.157}
\definecolor{grass}{rgb}{0.000,1.000,0.000}
\definecolor{vegetation}{rgb}{0.439,0.580,0.125}
\definecolor{tree}{rgb}{0.251,0.251,0.000}
\definecolor{groundobstacles}{rgb}{1.000,1.000,0.000}
\definecolor{person}{rgb}{1.000,0.063,1.000}
\definecolor{bicycle}{rgb}{1.000,0.800,0.600}
\definecolor{vehicle}{rgb}{0.000,0.502,0.502}
\definecolor{water}{rgb}{0.000,0.000,1.000}
\definecolor{building}{rgb}{1.000,0.000,0.000}
\definecolor{roof}{rgb}{0.251,0.627,0.471}
\definecolor{cables}{rgb}{1.000,0.627,0.000}
\definecolor{cabletower}{rgb}{0.416,0.000,1.000}
\definecolor{flyinganimals}{rgb}{0.961,0.592,0.412}
\definecolor{truck}{rgb}{0.500,0.500,0.250}
\definecolor{parkinglot}{rgb}{0.502,0.251,0.502}
\definecolor{constructions}{rgb}{0.941,0.471,0.471}
\definecolor{cranes}{rgb}{1.000,1.000,0.502}
\makeatletter
\newcommand{\road@occuflyfreq}{0.9377}
\newcommand{\walkway@occuflyfreq}{1.0560}
\newcommand{\dirt@occuflyfreq}{1.2102}
\newcommand{\gravel@occuflyfreq}{0.7960}
\newcommand{\rock@occuflyfreq}{0.0210}
\newcommand{\grass@occuflyfreq}{4.1978}
\newcommand{\vegetation@occuflyfreq}{2.3234}
\newcommand{\tree@occuflyfreq}{6.8357}
\newcommand{\groundobstacles@occuflyfreq}{1.7566}
\newcommand{\person@occuflyfreq}{0.0002}
\newcommand{\bicycle@occuflyfreq}{0.0046}
\newcommand{\vehicle@occuflyfreq}{0.5195}
\newcommand{\water@occuflyfreq}{1.1322}
\newcommand{\building@occuflyfreq}{75.7457}
\newcommand{\roof@occuflyfreq}{1.7417}
\newcommand{\cables@occuflyfreq}{0.0017}
\newcommand{\cabletower@occuflyfreq}{0.0040}
\newcommand{\flyinganimals@occuflyfreq}{0.0001}
\newcommand{\parkinglot@occuflyfreq}{1.4349}
\newcommand{\constructions@occuflyfreq}{0.1689}
\newcommand{\cranes@occuflyfreq}{0.0052}
\newcommand{\truck@occuflyfreq}{0.1067}
\newcommand{\occuflyfreq}[1]{{\csname #1@occuflyfreq\endcsname}}
\makeatother
\newcommand{\classfreq}[1]{{~\tiny(\occuflyfreq{#1}\%)}}

\usepackage{cuted}
\usepackage{multirow}
\usepackage[group-separator={,}, output-decimal-marker={.}, group-minimum-digits=4, group-digits=integer]{siunitx}
\title{OccuFly: A 3D Vision Benchmark for\\Semantic Scene Completion from the Aerial Perspective}
\author{Markus Gross\textsuperscript{1,2,3,*} \qquad Sai B. Matha \textsuperscript{1} \qquad Aya Fahmy\textsuperscript{1}\\
Rui Song\textsuperscript{4} \qquad Daniel Cremers\textsuperscript{2,3} \qquad Henri Meeß\textsuperscript{1}\\
\textsuperscript{1}Fraunhofer Institute IVI \quad \textsuperscript{2}TU Munich \quad \textsuperscript{3}MCML \quad \textsuperscript{4}UCLA
}

\begin{document}

\maketitle
\begin{strip}
  \vspace{-1.6cm}
  \centering
  \captionsetup{type=figure}
\includegraphics[trim={0cm 0cm 0cm 0cm},clip,width=\linewidth]{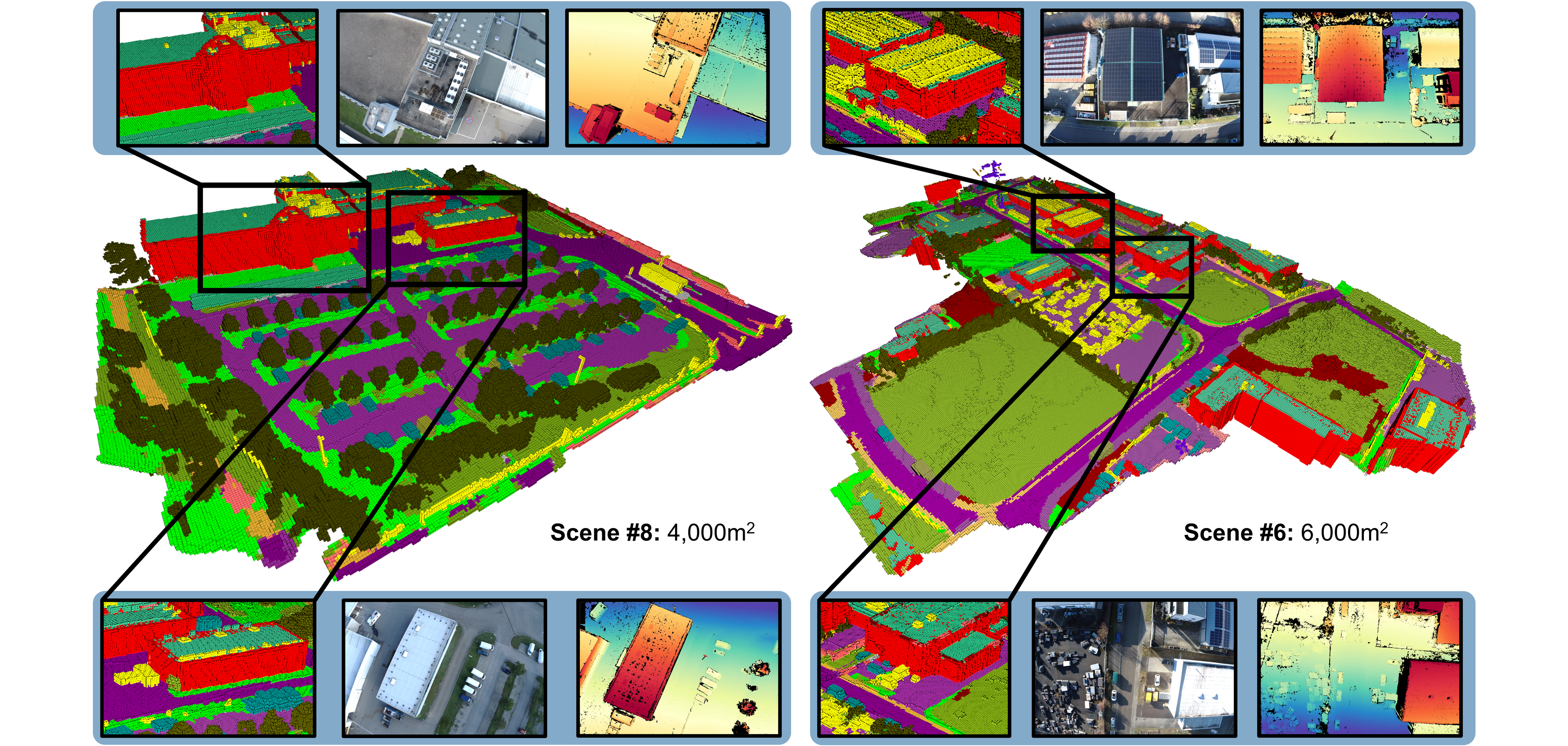}
  \captionof{figure}{OccuFly introduces the first real-world, aerial 3D SSC benchmark dataset, consisting of $9$ scenes with over \SI{20000}{} samples of RGB images, semantic occupancy grids, and metric depth maps, including $21$ semantic classes. OccuFly covers almost \SI{200000}{m\textsuperscript{2}} at \SI{50}{m}, \SI{40}{m}, and \SI{30}{m} altitude in urban, industrial, and rural environments during spring, summer, fall, and winter. Zoom in for the best view.}
  \label{fig_banner}
\end{strip}

\begingroup
\renewcommand\thefootnote{}
\footnotetext{%
\parbox[t]{\linewidth}{%
\textsuperscript{*}Corresponding author: markus.gross@tum.de
}}
\endgroup

\begin{abstract}
Semantic Scene Completion (SSC) is essential for 3D perception in mobile robotics, as it enables holistic scene understanding by jointly estimating dense volumetric occupancy and per-voxel semantics.
Although SSC has been widely studied in terrestrial domains such as autonomous driving, aerial settings like autonomous flying remain largely unexplored, thereby limiting progress on downstream applications.
Furthermore, LiDAR sensors are the primary modality for SSC data generation, which poses challenges for most uncrewed aerial vehicles (UAVs) due to flight regulations, mass and energy constraints, and the sparsity of LiDAR point clouds from elevated viewpoints.
To address these limitations, we propose a LiDAR-free, camera-based data generation framework.
By leveraging classical 3D reconstruction, our framework automates semantic label transfer by lifting <10\% of annotated images into the reconstructed point cloud, substantially minimizing manual 3D annotation effort.
Based on this framework, we introduce OccuFly, the first real-world, camera-based aerial SSC benchmark, captured across multiple altitudes and all seasons.
OccuFly provides over 20,000 samples of images, semantic voxel grids, and metric depth maps across 21 semantic classes in urban, industrial, and rural environments, and follows established data organization for seamless integration.
We benchmark both SSC and metric monocular depth estimation on OccuFly, revealing fundamental limitations of current vision foundation models in aerial settings and establishing new challenges for robust 3D scene understanding in the aerial domain. Visit \url{https://github.com/markus-42/occufly}.
\end{abstract}
    
\section{Introduction}
\label{sec_introduction}
Modern approaches in 3D computer vision enable holistic scene understanding for downstream applications such as autonomous navigation, surveillance, and augmented reality~\cite{dourado2021edgenet}.
To this end, one essential approach is Semantic Scene Completion (SSC)~\cite{monoscene}, which jointly infers the complete geometry of a 3D scene from a sparse observation, such as a camera image or LiDAR scan, while simultaneously assigning semantic labels to each element, typically represented in a voxelized 3D occupancy grid~\cite{thrun2005probabilistic}. Recently, SSC has been extended to Panoptic Scene Completion by incorporating instance-level awareness \cite{gross2025ipformerNeurIPS}.
While strong industry funding has led the research community to focus on terrestrial environments such as autonomous driving,
low-altitude aerial scene understanding for autonomous flying of uncrewed aerial vehicles (UAVs) is largely restricted to either 2D datasets~\cite{cai2025vdd, rizzoli2023syndrone, Fonder2019MidAir, kolbeinsson2024ddos, Chen2020valid, marcu2023dronescapes, lyu2020uavid, chen2018udd, azimi2019skyscapes, nigam2018aeroscapes, icgDataset, zhu2021visdrone, Wu2023uavd4l, yan2021crossloc, wang2025griffin, tian2025UCDNet, wang2024DHD}, or 3D mesh and point cloud datasets~\cite{dhao2025DSC3D, li2023matrixcity, beche2025claravid, espadaDataset, liqiang2022UrbanScene3D, wang2020tartanair, vuong2025aerialmegadepth, ye2024uav3d}.
Notably, the SSC objective remains largely unexplored from the aerial perspective, as no dedicated real or synthetic datasets exist, thereby confining related downstream applications to terrestrial environments.

Technically, SSC datasets are typically generated by (i) fusing multiple sparse LiDAR sweeps with registered poses to capture occluded regions as a dense point cloud, where (ii) each point is manually annotated with semantic labels and (iii) subsequently voxelized to produce SSC ground-truth \cite{zhang2024vision}.
While effective for ground vehicles, such LiDAR-based data generation becomes challenging in aerial environments.
First, multi-modal sensor setups for UAVs are not as widespread or advanced as in autonomous driving, since UAV platforms are subject to strict flight regulations, such as US~\cite{faa2016} or EU~\cite{easa2019} regulations, and they must adhere to stringent mass and energy constraints, which conflict with the heavier and more power-demanding nature of LiDARs compared to cameras.
Second, LiDAR sparsity persists and may even worsen from an elevated vantage point, leaving significant areas unobserved and unlabeled, which in turn would yield incomplete or low-quality ground-truth.

To address these limitations, we introduce OccuFly, the first real-world, low-altitude 3D vision benchmark for aerial Semantic Scene Completion.
Crucially, we propose a data generation framework that is based on camera modality, which is considered to be ubiquitous on modern UAVs.
Our dataset provides over 20,000 samples, resulting in 5$\times$ the number of samples and 6$\times$ the number of voxels compared to SemanticKITTI~\cite{behle2019semanticKitti}, which introduced the first SSC dataset for autonomous driving. 
Furthermore, we evaluate the state-of-the-art on OccuFly, yielding a comprehensive 3D aerial vision benchmark.\newline

Our \textbf{contributions} are summarized as follows:
\begin{itemize}
    \item We propose a novel and scalable data generation framework to construct SSC ground-truth, thereby (i) relying on camera modality to avoid LiDAR-based point cloud sparsity, (ii) avoiding LiDAR hardware to adhere to mass and energy constraints of most UAVs,  and (iii) reducing manual semantic labeling from tedious 3D annotation to efficient 2D annotation.
    \item Based on this framework, we present OccuFly, the first real-world aerial SSC benchmark consisting of $9$ scenes that provide over \SI{20000}{} samples of nadir and oblique perspective images with corresponding 3D semantic voxel grids, including $21$ semantic classes. OccuFly covers almost \SI{200000}{m\textsuperscript{2}} at \SI{50}{m}, \SI{40}{m}, and \SI{30}{m} altitude in urban, industrial, and rural environments across all seasons.
    \item In addition to the SSC samples, OccuFly offers more than \SI{20000}{} per-frame metric depth maps, on which we further train and release the Depth-Anything-v2~\cite{yang2024depthAnythingV2} depth estimation method, enabling state-of-the-art SSC.
    \item We identify a consistent domain gap in which state-of-the-art (i) terrestrial SSC models, (ii) 2D vision foundation models, and (iii) 3D vision foundation models fail to generalize to aerial data, thereby positioning OccuFly as a valuable and robust dataset to address these limitations.
\end{itemize}

\section{Related Work}
\label{sec_related_work}

\begin{figure*}[t!]
	\centering
	\includegraphics[width=.99\textwidth]{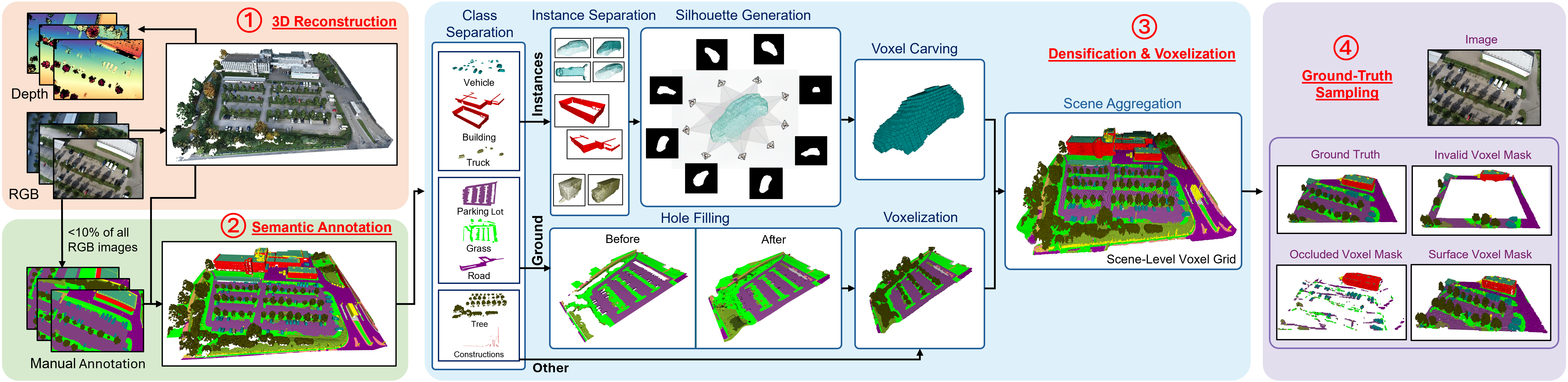}
	\caption{Proposed image-based data generation framework. An overview is provided in \cref{subsec_overview}. Zoom in for the best view.}
	\label{fig_data_gen_framework}    
\end{figure*}

\textbf{Datasets and Benchmarks.}\quad
Apart from indoor SSC~\cite{song2017indoorSSC}, the first outdoor SSC dataset was proposed by SemanticKITTI~\cite{behle2019semanticKitti} for autonomous driving.
Despite its impact, the limited scale and diversity of SemanticKITTI impeded the development of generalizable SSC models and their comprehensive evaluation~\cite{li2024sscbench}.
To address these limitations, multiple impactful datasets followed, such as nuScenes~\cite{caesar2020nuscenes}, Waymo~\cite{sun2020waymoOpenDataset}, and KITTI-360~\cite{liao2023kitti-360}, all relying on LiDAR modality for data generation.
As elaborated in \cref{sec_introduction}, LiDAR-based SSC ground-truth generation faces fundamental challenges stemming from sparse point clouds, as many regions remain unobserved and consequently unlabeled. Occlusions, misalignment from aggregating multiple sweeps, and dynamic objects further exacerbate these gaps, creating inaccuracies in the resulting volumetric labels. Additionally, the process of manually annotating such sparse 3D point clouds is both time-consuming and error-prone, undermining scalability for large-scale 3D reconstruction tasks.

To mitigate these challenges, several benchmarks have emerged.
Specifically, Occ3D~\cite{tian2023occ3d} combines multi-sweep densification with mesh reconstruction and camera filtering to reduce occlusions and label noise.
Meanwhile, OpenOccupancy~\cite{wang2023openoccupancy} combines pseudo-label generation with extensive human annotation to double the density of occupancy labels and refine boundaries.
In parallel, OCFBench~\cite{Liu2023ocfbench} synchronizes dynamic objects to address spatiotemporal occlusions, and excludes unknown voxels through ray-casting, thereby mitigating LiDAR sparsity.
Similarly, SSCBench~\cite{li2024sscbench} excludes unobserved voxels from training and evaluation, and integrates data from multiple sources to enhance geographic diversity.
Additionally, OpenScene~\cite{openscene2023} leverages voxel densification and flow integration to fill unobserved regions.
Finally, UniOcc~\cite{wang2025uniocc} unifies real and synthetic (CARLA~\cite{song2024cohff}, OpenCOOD~\cite{xu2022opv2v}) datasets, 
and provides label-free evaluation metrics to address suboptimal LiDAR coverage.
Notably, these benchmarks obtain depth maps implicitly by projecting LiDAR points onto images, relying on precise sensor extrinsics that can introduce geometric inaccuracies.

In contrast, our proposed data generation framework reconstructs metric 3D point clouds from geo-referenced imagery. 
We unproject manually annotated 2D semantic masks into these reconstructions via 2D–3D correspondences, yielding accurate per-point semantics, which substantially minimizes 3D annotation effort.
We further refine and densify the semantic point cloud, voxelize it, and extract consistent per-frame voxel grids and metric depth maps.\\

\textbf{Methods.}\quad Apart from recurrent~\cite{wang2025MonoMRN} and multi-view SSC methods~\cite{multiViewSSC_1, song2024cohff, wei2023surroundocc,guo2025sgformer}, single-view SSC was pioneered by MonoScene~\cite{monoscene}, which bridges 2D and 3D representations through an optics-inspired feature projection and a 3D context relation prior.
Recent advancements have explored diverse feature representations to bridge 2D image cues with volumetric representations, broadly categorized into planar and voxel-based approaches \cite{xu2025survey}.

In particular, planar representations, such as bird's-eye-view (BEV) \cite{li2022bevformer} or tri-perspective views (TPV), enable compact feature aggregation and long-range context modeling. TPVFormer~\cite{huang2023tpvformer} introduces the TPV framework with perpendicular planes and a transformer encoder for lifting multi-view features into semantic occupancy grids, while DISC~\cite{liu2025disc} extends BEV with discriminative queries and dual-attention decoders to disentangle instance-scene contexts for category-specific interactions. Building on hybrid designs, CGFormer~\cite{cgformer} fuses voxel and TPV spaces via context-aware queries and 3D deformable cross-attention, enhancing depth-aware aggregation forsuperior fidelity.

In contrast, voxel-based methods directly operate on 3D grids for fine-grained occupancy and semantics. Two-stage transformers like VoxFormer~\cite{li2023voxformer} and VisHall3D~\cite{lu2025vishall3d} employ sparse queries from depth priors followed by densification or visibility-aware decoupling to handle visible and occluded regions, respectively. Symphonies~\cite{symphonies} refines voxel queries with class-centric instance propagation for dynamic 2D-3D reasoning, whereas SOAP~\cite{soap} incorporates adaptive decoders drawing from semantic repositories.
Notably, many of these methods~\cite{li2023voxformer, symphonies, cgformer, liu2025disc} apply depth estimation to initialize voxel proposals as geometric priors.

Specialized enhancements include tri-axis approaches~\cite{bae2025scanssc}, State Space Models \cite{Li_2025_OccMamba,li2025stateSpaceSSC}, self-supervised methods~\cite{hayler2024s4c,Jevtic2025SceneDINO}, and Gaussian representations~\cite{haung2025gaussianFormer2}.
We refer the reader to the surveys of \cite{xu2025survey, zhang2024vision} for a comprehensive discussion.

\section{Data Generation Framework}
\label{sec_methodology}

\subsection{Overview}
\label{subsec_overview}

Our proposed data generation framework consists of four modules, shown in \cref{fig_data_gen_framework}. We utilize geo-referenced images to apply traditional multi-view reconstruction, generating a metric 3D point cloud (\cref{subsubsec_3d_reconstruction}).
This approach additionally yields metric depth maps and 2D–3D correspondences, allowing image pixels to be associated with reconstructed 3D points, effectively streamlining the creation of 3D semantic annotations.
Technically, we enable highly efficient label transfer by manually annotating only a small subset of the camera images (\SI{<10}{\%} on average) and lifting the semantic pixels into the point cloud (\cref{subsubsec_semantic_annotation}).
This reduces costly 3D annotation to efficient 2D image labeling, substantially lowering annotation effort.
Subsequently, we densify individual objects with our novel densification pipeline and eventually voxelize the semantic point cloud (\cref{subsubsec_voxelization}).
As all previous steps are performed on a global scene level, we finally retrieve per-frame ground-truth grids by frustum-culling the scene voxel grid using geo-referenced camera poses and intrinsics, resulting in one fixed-size semantic voxel grid per camera frame.
Similar to SemanticKITTI \cite{behley2019semantickitti}, we additionally construct binary masks to distinguish surface, occluded, and invalid voxels (\cref{subsubsec_ground_truth_sampling}).

\subsection{Problem Formulation}
\label{subsec_problem_formulation}

Given a set of $N$ geo-referenced, calibrated RGB images \mbox{$\mathcal{I}=\{\mathbf{I}_n \in \mathbb{R}^{H\times W\times 3}\}_{n=1}^{N}$}, acquired by a pinhole camera with per-frame intrinsics $\mathcal{K}=\{\mathbf{K}_n \in \mathbb{R}^{3\times 3}\}_{n=1}^{N}$, and world-to-camera poses $\mathcal{T}=\{\mathbf{T}^{\,n}_{c\leftarrow w} \in \text{SE}(3)\}_{n=1}^{N}$, we adopt a fixed-size metric voxel-grid specification $(X,Y,Z,r)$, where $X,Y,Z$ denote the number of voxels along cartesian axes, with voxel edge length \mbox{$r>0$}.
Moreover, we define a semantic label set \mbox{$\mathcal{C}=\{1,2,\dots,C\}$}.
Each image $\mathbf{I}_n$ is coupled to a ground-truth semantic voxel grid \mbox{$\mathbf{Y}_n \in \mathcal{C}^{X\times Y\times Z}$}, and a metric depth map \mbox{$\mathbf{D}_n\in\mathbb{R}^{H\times W}$}, yielding the dataset as image–grid-depth samples $\{(\mathbf{I}_n,\mathbf{Y}_n,\mathbf{D}_n)\}_{n=1}^{N}$, organized into multiple scenes with scene-dependent sample counts.

\subsection{Method}
 \label{subsec_method}

\subsubsection{3D Reconstruction}
\label{subsubsec_3d_reconstruction}

From the calibrated camera intrinsics $\mathcal{K}$, and the geo-referenced images $\mathcal{I}$ and their poses $\mathcal{T}$, we obtain a metric scene reconstruction via Structure-from-Motion (SfM)~\cite{schoenberger2016sfm} and Multi-View Stereo (MVS)~\cite{schoenberger2016mvs}, abstracted as
\begin{equation}
(\mathcal{P},\,\mathcal{D},\,\mathcal{A})\;=\;\Psi_{\text{SfM+MVS}}(\mathcal{I},\mathcal{K},\mathcal{T}),
\end{equation}
where the set $\mathcal{P}=\{\mathbf{x}_m\in\mathbb{R}^3\}_{m=1}^{M}$ represents a dense point cloud with $M=|\mathcal{P}|$ number of points in world coordinates, and \mbox{$\mathcal{D}=\{\mathbf{D}_n\in\mathbb{R}^{H\times W}\}_{n=1}^{N}$} are per-image metric depth maps. Moreover, $\mathcal{A}_n$ denotes per-image 2D–3D correspondences , which satisfy the projection of a 3D point $\mathbf{x}$ onto a 2D image pixel $(u,v)$, formulated as
\begin{equation}
    \mathcal{A}_n=\big\{\, ((u,v),\mathbf{x}) \,\big|\, (u,v,1)^\top\sim \mathbf{K}_n[\,\mathbf{R}_n\,|\,\mathbf{t}_n\,][\mathbf{x}^\top\ 1]^\top\big\},
\end{equation}
where $\sim$ denotes equality up to a non-zero scalar, and $\mathbf{R}_n$ and $\mathbf{t}_n$ are the rotational and translational components of $\mathbf{T}^{\,n}_{c\leftarrow w}$, respectively.
All correspondences of a scene are given by $\mathcal{A}=\bigcup_{n=1}^{N}\mathcal{A}_n$.

\subsubsection{Semantic Annotation}
\label{subsubsec_semantic_annotation}

We define a finite, non-empty set of semantic classes $\mathcal{C}=\{1,\dots,C\}$.
To minimize manual annotation effort, we annotate only a small subset $\mathcal{J}\subset\{1,\dots,N\}$ of images, exploiting the fact that each 3D point is observed by multiple cameras.
Note that $\mathcal{J}$ denotes the index set of images, not the images themselves.
We select $\mathcal{J}$ via spatially stratified sampling across the scene.
More specifically, we partition the ground plane into a regular grid of square cells with scene-dependent edge lengths of $23$ to $28$ meters.
For each cell center, we select the image whose pose is closest to that location. Additionally, for cells at the scene border, we select the pose closest to the border.
Using the 2D--3D correspondences $\mathcal{A}$ (see \cref{subsubsec_3d_reconstruction}), we quantify the coverage of reconstructed points by
\small
\begin{equation}
\rho(\mathcal{J}) \;=\; \frac{\big|\{\,\mathbf{x}\in\mathcal{P} \,\big|\, \exists\,n\in\mathcal{\mathcal{J}},\,\exists\,(u,v)\ \text{s.t.}\ ((u,v),\mathbf{x})\in\mathcal{A}_n\,\}\big|}{|\mathcal{P}|}
\label{eq_reconstruction_coverage}
\end{equation}
\normalsize
Empirically, \mbox{$\rho(\mathcal{J})>0.99$}, while $|\mathcal{J}|/N< 0.1$ on average, implying that annotating less than $10\%$ of all images covers more than $99\%$ of all 3D points, further detailed in \cref{subsec_dataset_evaluation}. 

After manually annotating all images in $\mathcal{J}$, we lift semantic labels from images to points via back-projection.
Note that this step is similar to~\cite{yan2024pointssc}, but adapted to a LiDAR-free, camera-only aerial setting based on SfM+MVS reconstruction.
As most of the points are observed by multiple cameras, we fuse multi-view evidence to assign robust per-point labels.
Technically, this is accomplished by unweighted majority voting, where ties are broken by a fixed class-prior order derived from class frequencies.
Furthermore, to annotate unlabeled points, we apply k-nearest-neighbor (kNN) with inverse-distance weights within a fixed neighborhood. As a denoising step, we apply a second iteration of kNN to all points, effectively relabeling every point to the dominant class in its neighborhood, similar to~\cite{tian2023occ3d}.
The resulting semantic point cloud $\mathcal{P}_{\mathcal{S}}=\{(\mathbf{x}_m,c_m)\}_{m=1}^{M}$ consists of 3D points $\mathbf{x}_m$ and corresponding semantic class labels $c_m \in \mathcal{C}$.

\subsubsection{Class-Aware Densification and Voxelization}
\label{subsubsec_voxelization}

\paragraph{Preliminaries.}
Given the semantic point cloud $\mathcal{P}_{\mathcal{S}}=\{(\mathbf{x}_m,c_m)\}_{m=1}^{M}$ from \cref{subsubsec_semantic_annotation}, we first partition the semantic classes into three disjoint groups
\begin{equation}
\mathcal{C}_{\text{inst}}\;\cup\;\mathcal{C}_{\text{gnd}}\;\cup\;\mathcal{C}_{\text{oth}} \;=\; \mathcal{C},\qquad
\mathcal{C}_{\text{inst}}\cap\mathcal{C}_{\text{gnd}}\cap\mathcal{C}_{\text{oth}}=\varnothing,
\end{equation}
corresponding to instance classes $\mathcal{C}_{\text{inst}}$ that are to be densified object-wise (\eg, vehicles), ground classes $\mathcal{C}_{\text{gnd}}$ that are to be surface-reconstructed (\eg, road), and other classes $\mathcal{C}_{\text{oth}}$ that are directly voxelized (\eg, constructions). Based on this class partition, we apply \textbf{group separation} to retrieve group-specific point cloud subsets
\begin{align}
    \mathcal{P}_{\text{inst}}&=\{(\mathbf{x},c)\in\mathcal{P}_{\mathcal{S}} \,|\, c\in\mathcal{C}_{\text{inst}}\}\,,\\
    \mathcal{P}_{\text{gnd}}&=\{(\mathbf{x},c)\in\mathcal{P}_{\mathcal{S}} \,|\, c\in\mathcal{C}_{\text{gnd}}\}\,,\\
    \mathcal{P}_{\text{oth}}&=\{(\mathbf{x},c)\in\mathcal{P}_{\mathcal{S}} \,|\, c\in\mathcal{C}_{\text{oth}}\}\,.
\end{align}

Furthermore, let $\mathcal{G} = [X] \times [Y] \times [Z]$ be a voxel grid. For any point cloud subset $\mathcal{Q}\subset\mathcal{P}$ and a target voxel resolution $r>0$, we denote by $\text{Vox}_r(\mathcal{Q})\subset\mathcal{G}$ the set of occupied voxels obtained by standard binning (point rasterization)~\cite{Faugeras1993pointCloudVoxelization} at resolution $r$. For triangle meshes, we use the same notation to indicate triangle-to-voxel scan conversion~\cite{Kaufman1986meshVoxelization}.

\paragraph{Instance classes.}

We apply \textbf{instance separation} by decomposing $\mathcal{P}_{\text{inst}}$ into object instances using Euclidean clustering via DBSCAN~\cite{ester1996_dbscan}
with class-specific parameters $(\varepsilon_c,\text{minPts}_c)$. This process yields $J$ object instances
\begin{equation}
\mathbb{S}=\text{DBSCAN}(\mathcal{P}_{\text{inst}},\varepsilon_c,\text{minPts}_c)=\{\mathcal{S}_j\subset\mathcal{P}_{\text{inst}}\}_{j=1}^{J},
\end{equation}
where $\mathcal{S}_j$ represents an instance point cloud.

To create the voxelized visual hull from an instance point cloud $\mathcal{S}$, we perform two major steps: Silhouette extraction and silhouette-based voxel carving. 
For each instance $\mathcal{S}\in\mathbb{S}$, let $\mathrm{pos}(\mathcal{S})=\{\,\mathbf{x}\,|\,(\mathbf{x},c)\in\mathcal{S}\,\}$ denote its 3D positions.

To \textbf{extract silhouettes}, we (i) place $K$ virtual cameras $\mathbb{V}=\{(\mathbf{K}_k,\mathbf{T}^{\,k}_{c\leftarrow w})\}_{k=1}^{K}$ quasi-uniformly distributed on the viewing sphere around $\mathrm{pos}(\mathcal{S})$ (in practice we use $K{=}24$);
(ii) project $\mathrm{pos}(\mathcal{S})$ to each view to obtain 2D point sets $\mathcal{U}_k=\{\,\pi_k(\mathbf{x}) \,|\, \mathbf{x}\in\mathrm{pos}(\mathcal{S})\,\}$, where $\pi_k$ is the pinhole projection induced by $(\mathbf{K}_k,\mathbf{T}^{\,k}_{c\leftarrow w})$;
and (iii) compute a binary silhouette $\Omega_k\subset\mathbb{R}^2$ via the $\alpha$-shape boundary~\cite{edelsbrunner1983ashape} of $\mathcal{U}_k$, where $\Omega_k$ is the set of pixels that belong to the instance.

For \textbf{voxel carving}, we apply multi-view silhouette carving~\cite{273735}. To this end, we back-project each silhouette to a generalized cone
\begin{equation}
\mathcal{R}_k \;=\; \{\,\mathbf{x}\in\mathbb{R}^3 : \pi_k(\mathbf{x})\in\Omega_k \,\}.
\end{equation}
The continuous instance hull is $\mathcal{H}(\mathcal{S})=\bigcap_{k=1}^{K}\mathcal{R}_k$. To carve within a finite space around the instance hull, let $\mathcal{B}(\mathcal{S})$ be a tight axis-aligned bounding box of $\mathrm{pos}(\mathcal{S})$, dilated by a small margin.
We discretize $\mathcal{B}(\mathcal{S})$ into a 3D grid of voxels indexed by $\mathbf{v} = (i, j, k)$.
The carved occupancy set is
\begin{equation}
\mathcal{O}_{\text{inst}}(\mathcal{S}) = \left\{ \mathbf{v} \in \mathcal{G} \,|\, \text{center}(\mathbf{v}) \in \mathcal{B}(\mathcal{S}) \cap \mathcal{H}(\mathcal{S}) \right\}.
\end{equation}
All voxels $\mathbf{v}\in\mathcal{O}_{\text{inst}}(\mathcal{S})$ receive the semantic label of its original semantic instance point cloud $\mathcal{P}_{\text{inst}}$. Finally, we aggregate all instances via $\mathcal{O}_{\text{inst}} = \bigcup_{\mathcal{S} \in \mathbb{S}} \mathcal{O}_{\text{inst}}(\mathcal{S})$.

\paragraph{Ground classes.}
We densify $\mathcal{P}_{\text{gnd}}$ via Poisson surface reconstruction~\cite{poisson} to obtain a watertight triangle mesh $\mathcal{M}_{\text{gnd}}=\Psi_{\text{Poisson}}(\mathcal{P}_{\text{gnd}})$, which \textbf{fills holes} and enforces surface continuity, similar to \cite{wei2023surroundocc}. We then voxelize the mesh:
\begin{equation}
\mathcal{O}_{\text{gnd}} \;=\; \text{Vox}_r(\mathcal{M}_{\text{gnd}}),
\end{equation}
assigning to each occupied voxel the majority ground class of contributing mesh samples in its cell.

\paragraph{Other classes.}
For $\mathcal{P}_{\text{oth}}$, we apply direct voxelization:
\begin{equation}
\mathcal{O}_{\text{oth}} \;=\; \text{Vox}_r(\mathcal{P}_{\text{oth}}),
\end{equation}
with per-voxel semantics determined by majority voting of points falling into the voxel.

\paragraph{Aggregation.}
We construct a scene-level semantic voxel grid by combining all groups with a fixed precedence order $\text{inst}\succ\text{oth}\succ\text{gnd}$ to resolve label conflicts:
\begin{equation}
\mathbf{Y}(\mathbf{v}) \;=\;
\begin{cases}
\text{label from }\mathcal{O}_{\text{inst}}, & \mathbf{v}\in \mathcal{O}_{\text{inst}},\\
\text{label from }\mathcal{O}_{\text{oth}},  & \mathbf{v}\in \mathcal{O}_{\text{oth}}\setminus \mathcal{O}_{\text{inst}},\\
\text{label from }\mathcal{O}_{\text{gnd}}, & \mathbf{v}\in \mathcal{O}_{\text{gnd}}\setminus (\mathcal{O}_{\text{inst}}\cup \mathcal{O}_{\text{oth}}),\\
0, & \text{otherwise,}
\end{cases}
\end{equation}
where $0$ denotes empty. The resulting $\mathbf{Y}$ constitutes the semantic voxel grid for the whole scene.

\subsubsection{Ground-Truth Sampling}
\label{subsubsec_ground_truth_sampling}

\paragraph{Frustum Culling.}Given the scene-level semantic voxel grid $\mathbf{Y}\in(\{0\}\cup\mathcal{C})^{X\times Y\times Z}$ and per-frame camera parameters $(\mathbf{K}_n,\mathbf{T}^{\,n}_{c\leftarrow w})$, we construct per-frame ground-truth by frustum-culling and rasterization at a fixed metric specification $(X,Y,Z,r)$. Let $\pi_n$ denote the pinhole projection induced by $(\mathbf{K}_n,\mathbf{T}^{\,n}_{c\leftarrow w})$, and let $[d_{\min},d_{\max}]$ be near/far clipping distances, respectively. Define the truncated frustum
\[
\mathcal{F}_n \;=\; \big\{\, \mathbf{x}\in\mathbb{R}^3 \,:\, \pi_n(\mathbf{x})\in[0,W)\!\times\![0,H)\big\},
\]
with $ d_{\min}\le d_n(\mathbf{x})\le d_{\max}$, where $d_n(\mathbf{x})$ is the camera-centric depth of $\mathbf{x}$. We obtain the per-frame grid $\mathbf{Y}_n\in(\{0\}\cup\mathcal{C})^{X\times Y\times Z}$ by discretizing $\mathcal{F}_n$ at resolution $r$ and sampling $\mathbf{Y}$ at voxel centers:
\[
\mathbf{Y}_n(\mathbf{v}) \;=\; 
\begin{cases}
\mathbf{Y}(\mathbf{v}), & \text{if }\text{center}(\mathbf{v})\in \mathcal{F}_n,\\
0, & \text{otherwise,}
\end{cases}
\quad \mathbf{v}\in\mathcal{G}.
\]

\paragraph{Binary Masks.}In addition, we construct three binary masks, similar to SemanticKITTI~\cite{behley2019semantickitti}: invalid $\mathbf{M}^{\text{inv}}_n$, surface $\mathbf{M}^{\text{surf}}_n$, and occluded $\mathbf{M}^{\text{occ}}_n$, all in $\{0,1\}^{X\times Y\times Z}$.

\textit{Invalid Mask.} The invalid mask represents voxels outside the field of view:
\begin{equation}
\mathbf{M}^{\text{inv}}_n(\mathbf{v}) \;=\; \mathbf{1}\!\big[\,\text{center}(\mathbf{v})\notin \mathcal{F}_n\,\big],
\end{equation}
where $\mathbf{1}[\cdot]$ denotes the indicator (Iverson) function.

\textit{Surface Mask (view-independent).} Let $\mathcal{N}_6(\mathbf{v})$ be the 6-neighborhood in the grid, and let $\mathcal{E}(\mathbf{v})=\mathbf{1}[\mathbf{Y}_n(\mathbf{v})\neq 0]$ denote occupancy. We mark geometric boundary voxels by
\[
\mathbf{M}^{\text{surf}}_n(\mathbf{v}) \;=\; \mathbf{1}\!\Big[\, \mathcal{E}(\mathbf{v})=1 \;\wedge\; \exists\,\mathbf{u}\in\mathcal{N}_6(\mathbf{v})\,\big|\, \mathcal{E}(\mathbf{u})=0 \,\Big].
\]

\textit{Occluded Mask (view-dependent).} For each pixel $(u,v)\in[0,W)\!\times\![0,H)$, consider the ordered set of frustum voxels $\mathcal{V}_n(u,v)=\langle \mathbf{v}_1,\mathbf{v}_2,\dots\rangle$ traversed by the ray $\rho_n(u,v)$ from near to far. Let $i^\star=\min\{i:\mathcal{E}(\mathbf{v}_i)=1\}$ if such $i$ exists. Then we assign
\[
\mathbf{M}^{\text{occ}}_n(\mathbf{v}_j) \;=\; 
\begin{cases}
0, & j=i^\star,\\
1, & j>i^\star \text{ and } \mathcal{E}(\mathbf{v}_j)=1,\\
0, & \text{otherwise,}
\end{cases}
\]
for all $\mathbf{v}_j\in\mathcal{V}_n(u,v)$. We set $\mathbf{M}^{\text{occ}}_n(\mathbf{v})=0$ if $\mathbf{v}$ lies outside all rays or is invalid.

Finally, the per-frame ground-truth sample consists of $(\mathbf{I}_n,\mathbf{Y}_n,\mathbf{M}^{\text{inv}}_n,\mathbf{M}^{\text{surf}}_n,\mathbf{M}^{\text{occ}}_n)$, where invalid voxels are excluded from evaluation, surface voxels represent view-independent geometric boundaries, and occluded voxels capture view-dependent occupied regions behind the first visible surface along camera rays.

\begin{table}[t!]
\centering
\caption{OccuFly dataset statistics, discussed in \cref{subsec_dataset_statistics}.}
\resizebox{\columnwidth}{!}{
\begin{tabular}{c|c|c|c|ccc|c}
\toprule
\multirow{2}{*}{Scene} & \multirow{2}{*}{Season} & \multirow{2}{*}{Environment} & \multirow{2}{*}{Area [m\textsuperscript{2}]} & \multicolumn{4}{c}{Number of Samples}   \\
\cline{5-8}
& & & & \SI{50}{m} & \SI{40}{m} & \SI{30}{m} & Total\\
\midrule
\rowcolor[HTML]{E6E6E6}
\multicolumn{1}{c}{\textbf{Training}} & \multicolumn{6}{c}{} & $\mathbf{14,804}$ \\
$01$ & Winter & Rural  & $41,234$ &  $406$ & $469$ & $512$ & $1,387$ \\
$02$ & Winter & Urban  & $8,529$  &  $277$ & $468$ & $132$ & $877$ \\
$03$ & Spring & Urban  & $25,077$ & $871$ & $1,315$ & $1,427$ & $3,613$ \\
$04$ & Spring & Industrial & $55,579$ & $1,140$ & $1,258$ & $1,564$ & $3,962$ \\
$05$ & Summer & Rural  & $24,810$ & $1,006$ & $1,682$ & $2,277$ & $4,965$ \\
\midrule
\rowcolor[HTML]{E6E6E6}
\multicolumn{1}{c}{\textbf{Validation}} & \multicolumn{6}{c}{} & $\mathbf{1,965}$\\
$06$ & Spring & Urban  & $5,428$  & $366$ & $266$ & $327$ & $959$ \\
$07$ & Spring & Industrial & $5,802$ & $279$ & $384$ & $343$ & $1,006$ \\
\midrule
\rowcolor[HTML]{E6E6E6}
\multicolumn{1}{c}{\textbf{Test}} & \multicolumn{6}{c}{} & $\mathbf{3,842}$\\
$08$ & Fall   & Industrial & $4,314$  & $183$ & $316$ & $383$ & $882$ \\
$09$ & Spring & Urban  & $23,165$ & $1,304$ & $1,416$ & $240$ & $2,960$ \\
\midrule
\textbf{Total} & \multicolumn{1}{c|}{} & \multicolumn{1}{c|}{} & \multicolumn{1}{c|}{$\mathbf{193,938}$} & $\mathbf{5,832}$ & $\mathbf{7,574}$ & $\mathbf{7,205}$ & $\mathbf{20,611}$ \\
\bottomrule
\end{tabular}
}
\label{tab_occufly}
\end{table}

\section{OccuFly Dataset}
\label{sec_occufly_dataset}

\subsection{Data Collection}
\label{subsec_data_collection}
We utilize two DJI UAV platforms for photogrammetric data acquisition~\cite{photogrammetry}: The Phantom~4 RTK (P4)~\cite{dji2025phantom4rtk} and the DJI Mavic 3 Enterprise Series (M3-ES)~\cite{dji2025mavic3enterprise}, capturing images at 5472$\times$3648 and 4000$\times$3000 pixels, respectively.
Geo-referenced camera poses and orientations are recorded by each flight controller via onboard sensor fusion (GNSS, IMU, compass, and magnetometer)~\cite{dji2025sensorfusion}. 
We collect data from nine scenes in urban, industrial, and rural environments, spanning spring, summer, fall, and winter.
All data were collected within the broader geographic region of \mbox{Ingolstadt}, Germany.
For data acquisition, we executed automated double-grid flight patterns at altitudes of \SI{50}{m}, \SI{40}{m}, and \SI{30}{m}. These missions yielded ground sampling distances (GSD) of \SI{1.4}{}, \SI{1.1}{}, and \SI{0.8}{cm/pixel} with the P4 platform, and \SI{6.7}{}, \SI{5.4}{}, and \SI{4.1}{cm/pixel} with the M3-ES platform. The double-grid pattern provided, on average, \SI{67}{\percent} side and \SI{74}{\percent} forward image overlap. For oblique acquisitions, camera tilt angles were set to \(-75^\circ\) at \SI{50}{m} and \SI{40}{m}, and \(-70^\circ\) at \SI{30}{m}. Additionally, using the M3-ES UAV, we collected nadir (\(0^\circ\) tilt) imagery at all altitudes.
Further details are provided in the supplementary material.

\subsection{Dataset Statistics}
\label{subsec_dataset_statistics}

We summarize OccuFly in \cref{tab_occufly}. The dataset comprises 9 scenes and more than \SI{20000}{} annotated samples, each including (i) an RGB image, (ii) a semantic voxel grid, and (iii) a metric depth map.
Voxel grids are annotated with 21 semantic classes that are detailed in the supplementary, including class frequencies.
OccuFly spans approximately \SI{193938}{m^2} across altitudes of \SI{50}{m}, \SI{40}{m}, and \SI{30}{m}, covering urban, industrial, and rural environments in spring, summer, fall, and winter.
3D space is discretized into voxel grids of resolution $192 \times 128 \times 128$ with voxel size $r=\SI{0.5}{m}$.
Ground-truth at \SI{40}{m} and \SI{30}{m} is generated via frustum culling (\cref{subsubsec_ground_truth_sampling}) of the \SI{50}{m} scene-level semantic grid.
Moreover, the dataset follows the SSCbench~\cite{li2024sscbench} data organization to streamline integration for existing methods.
Finally, we report semantic group assignments (\cref{subsubsec_voxelization}) together with scene-wise depth distributions in the supplementary material.

\begin{table}[t!]
\centering
\caption{Comparison of terrestrial and aerial vision-based SSC benchmark datasets, detailed in \cref{subsec_dataset_evaluation}.}
\resizebox{\linewidth}{!}{
\begin{tabular}{l|c|c|c|c|c|c}
\toprule
 & Camera & \# of & Depth & Semantic & 
\multicolumn{2}{c}{~Grid Resolution} \\
\cline{6-7}
 & Views & Samples & Maps & Classes & $X\times Y\times Z$ & $r$ \\ 
\midrule
\rowcolor[HTML]{E6E6E6}
\multicolumn{7}{l}{\textbf{Terrestrial Benchmarks}} \\
\midrule
SemanticKITTI \cite{behley2019semantickitti} & single & $4,649$ & \textcolor{red}{\xmk} & 19 & $256\times256\times32$ & $0.20$ \\
\midrule
OpenOccupancy \cite{wang2023openoccupancy} & multi  & $34,149$ & \textcolor{red}{\xmk} & 16 & $512\times512\times40$ & $0.20$ \\
\midrule
SSCBench-Waymo \cite{li2024sscbench} & multi  & $19,985$   & \textcolor{red}{\xmk} & 14 & $256\times256\times32$ & $0.20$ \\ 
SSCBench-nuScenes \cite{li2024sscbench} & multi  & $34,078$   & \textcolor{red}{\xmk} & 16 & $256\times256\times32$ & $0.20$ \\ 
SSCBench-KITTI-360 \cite{li2024sscbench} & multi  & $12,865$   & \textcolor{red}{\xmk} & 19 & $256\times256\times32$ & $0.20$ \\
\midrule
Occ3D-Waymo \cite{tian2023occ3d} & multi  & $200,000$   & \textcolor{red}{\xmk} & 14 & $3200\times3200\times128$ & $0.05$ \\
Occ3D-nuScenes \cite{tian2023occ3d} & multi  & $40,000$   & \textcolor{red}{\xmk} & 16 & $200\times200\times16$ & $0.40$ \\
\midrule
\rowcolor[HTML]{E6E6E6}
\multicolumn{7}{l}{\textbf{Aerial Benchmarks}} \\
\midrule
OccuFly (ours)  & single  & $20,611$   & \textcolor{green}{\cmk} & $21$ & $192 \times 128 \times 128$ & $0.50$ \\
\bottomrule
\end{tabular}
}
\label{tab_benchmark_datasets}
\end{table}

\begin{table}[t]
\centering
\caption{Comparison of OccuFly depth maps to other real-world low-altitude aerial datasets containing metric depth maps (Sec.~\ref{subsec_dataset_evaluation}).}
\resizebox{0.85\linewidth}{!}{
\begin{tabular}{c|c|c|c|c}
\toprule
Dataset & \# Depth Maps & Density & Environments & Seasons \\
\midrule
WildUAV~\cite{florea2021wilduav} & $\sim$\,\SI{1500}{} & Dense & Rural & \begin{tabular}[c]{@{}c@{}}Summer,\\Autumn\end{tabular} \\
\hline
UseGeo~\cite{Nex2024UseGeo} & \SI{829}{} & Sparse & Urban & \begin{tabular}[c]{@{}c@{}}n.a.\end{tabular} \\
\hline
OccuFly (ours) & \SI{20611}{} & Dense & \begin{tabular}[c]{@{}c@{}}Rural,\\Urban,\\Industrial\end{tabular} & \begin{tabular}[c]{@{}c@{}}Spring,\\Summer,\\Autumn,\\Winter\end{tabular} \\
\bottomrule
\end{tabular}
}
\label{tab_depth_datasets}
\end{table}

\subsection{Dataset Evaluation}
\label{subsec_dataset_evaluation}

\textbf{SSC Ground-Truth.}
Since OccuFly provides the first real-world aerial SSC dataset, we compare it against established terrestrial datasets (\cref{tab_benchmark_datasets}). Similar to SemanticKITTI~\cite{behley2019semantickitti}, which introduced real-world SSC to autonomous driving, OccuFly introduces real-world SSC to the aerial domain, but at a substantially larger scale: the number of samples is more than five times higher, and the total number of labeled voxels is over six times larger than SemanticKITTI. OccuFly further provides the largest class taxonomy while adhering to the SSCBench-style data structure~\cite{li2024sscbench} for seamless integration.

\textbf{Depth Map Ground-Truth.}
In \cref{tab_depth_datasets}, we compare OccuFly to other real-world, low-altitude aerial datasets that include metric depth maps.
To the best of our knowledge, WildUAV~\cite{florea2021wilduav} and UseGeo~\cite{Nex2024UseGeo} are the only publicly available datasets of this kind. OccuFly is substantially larger, providing more than $13\times$ and more than $24\times$ as many metric depth maps, respectively, while spanning a broader range of environments and seasons. This positions OccuFly as the largest and most diverse publicly available low‑altitude metric depth estimation dataset to date, which further enables holistic vision-based 3D scene understanding, such as SSC.

\begin{figure}[t!]
	\centering
	\includegraphics[width=.99\linewidth]{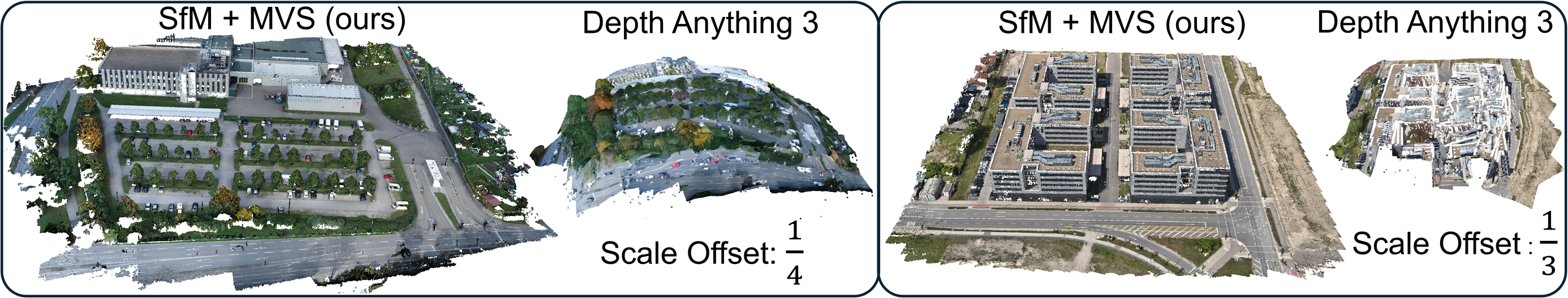}
	\caption{Evaluation of our classical 3D reconstruction compared to DepthAnything3~\cite{lin2025depthanything3} foundation model, detailed in \cref{subsec_dataset_evaluation}.}
	\label{fig_SfMMVS_vs_DA3}    
\end{figure}

\textbf{3D Reconstruction.}
We evaluate the 3D foundation model DepthAnything3~\cite{lin2025depthanything3} on all scenes of OccuFly, as illustrated in \cref{fig_SfMMVS_vs_DA3}, and observe an average metric scale deviation of \SI{526}{\percent}. 
This result supports our choice of classical SfM+MVS reconstruction and aligns with prior work~\cite{wu2025aerialblocks}, indicating that current 3D foundation models struggle to reliably handle aerial imagery.
Additionally, we quantify geometric consistency of the SfM+MVS pipeline using the standard RMSE reprojection error. Across all scenes, an average reprojection error of $1.24$ pixels demonstrates strong geometric fidelity~\cite{hartley2003multiple}, reflecting precise 2D-3D alignment under high-resolution inputs. Scene-wise reconstructions and reprojection errors are provided in the supplementary.

\textbf{Semantic Annotation Efficiency.}
As an alternative to manual annotation, we evaluate the use of 2D pseudo-labels following prior work such as PointSSC~\cite{yan2024pointssc}.
Using the 2D foundation model Semantic Segment Anything~\cite{chen2023semantic} to generate pseudo-labels yields poor segmentation quality, which we attribute to its backbone Segment Anything Model (SAM)~\cite{kirillov2023sam} that underperforms in the aerial domain~\cite{osco2023aerialSAM}.
To provide a stronger baseline, we train a ConvNeXt~\cite{liu2022convnext} model on a consolidation of seven established aerial segmentation datasets~\cite{nigam2018aeroscapes, rahn2021floodnet, icgDataset, lyu2020uavid, chen2018udd, cai2025vdd, speth2022otukama}. However, evaluation against our manual annotations results in only \SI{17.58}{\percent} mIoU, indicating that pseudo-labels remain highly unreliable in the aerial domain and reinforcing the need for accurate manual labeling.
To this end, our method leverages multi-view correspondences to lift only a small subset of manually annotated images to 3D.
Specifically, annotating fewer than 10\% of the images per scene suffices to automatically label over 99\% of reconstructed points, as determined by \cref{eq_reconstruction_coverage}, demonstrating a highly efficient and scalable annotation strategy.
We report scene-wise annotation ratios and coverage statistics in the supplementary material.

\textbf{Geometric Fidelity and Semantic Accuracy.}
Using RTK-based camera poses, we render the semantic point cloud into $135$ additional manual annotations across all scenes and altitudes that were not used during data generation.
This evaluation achieves \SI{92}{\percent} pixel-wise agreement,
confirming high geometric fidelity and semantic accuracy.

\begin{figure*}[ht!]
	\centering
	\includegraphics[width=.99\textwidth]{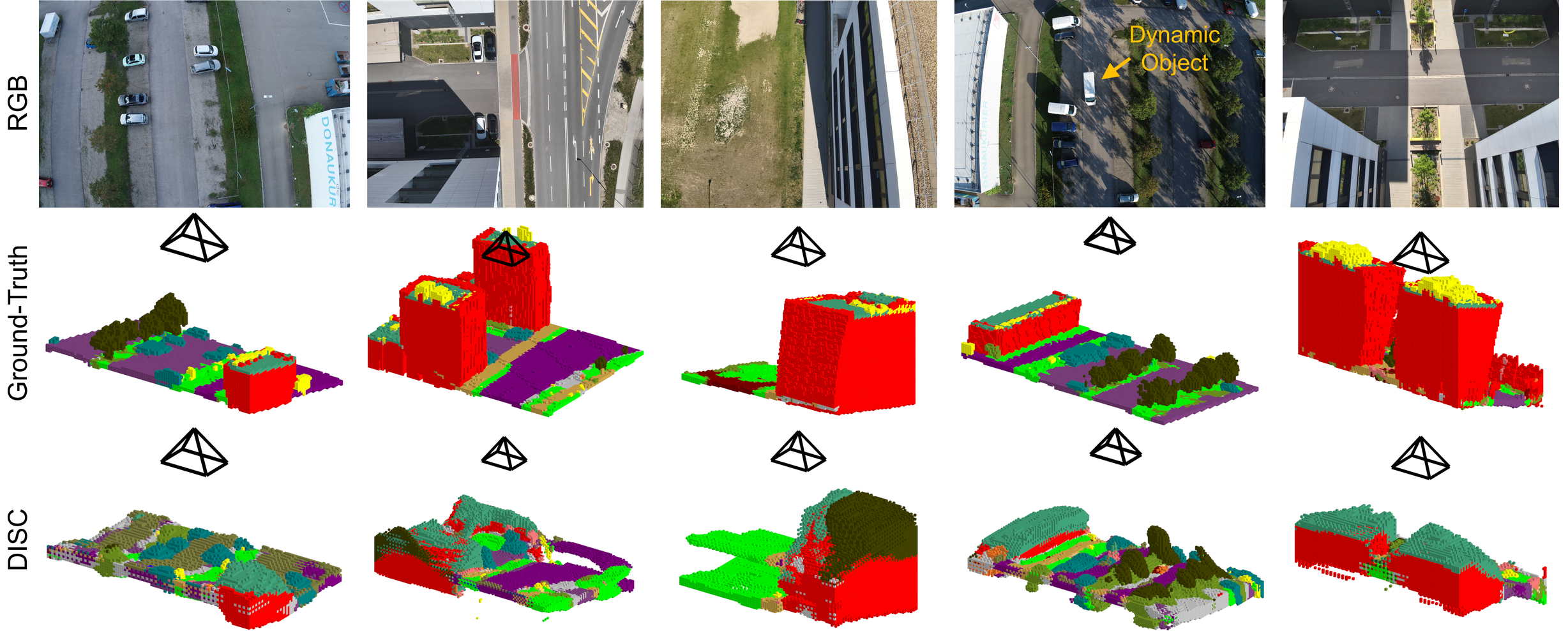}
	\caption{Visual SSC examples and qualitative SSC evaluation for DISC~\cite{liu2025disc} on the OccuFly test set, discussed in \cref{subsec_qualtitative_results}.}
	\label{fig_qualitative_results}    
\end{figure*}

\begin{figure}[t!]
	\centering
	\includegraphics[width=.99\linewidth]{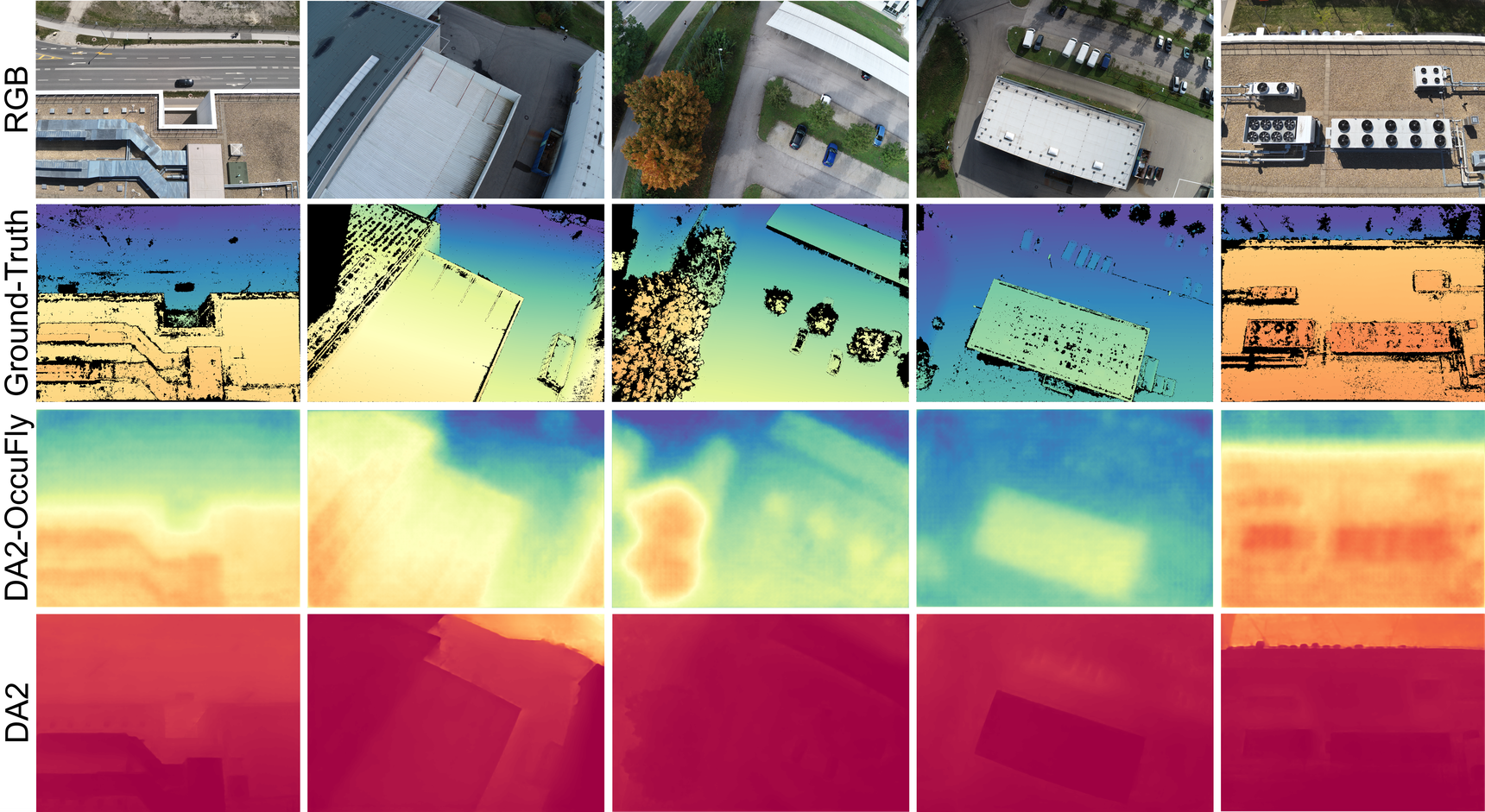}
	\caption{Qualitative evaluation of zero-shot and fine-tuned depth estimation for DepthAnything2~\cite{yang2024depthAnythingV2} on OccuFly test set (\cref{subsec_qualtitative_results}).}
	\label{fig_qualitative_depth}    
\end{figure}

\section{Benchmark Experiments}
\label{sec_experiments}

\subsection{Experimental Setup}
\label{subsec_experimental_setup}

\textbf{Aerial Semantic Scene Completion.}\quad 
We benchmark the state-of-the-art SSC methods Symphonies~\cite{symphonies} and DISC~\cite{liu2025disc}, using official implementations and evaluation protocols to ensure scientific rigor. 
As these methods require depth maps to back-project geometric priors (\cref{sec_related_work}), we replace their terrestrial depth estimator with our DepthAnything2-Occufly model (discussed in the next paragraph).
Further details are provided in the supplementary.\\
\noindent\textbf{Metric Monocular Depth Estimation.}\quad
This task is crucial for vision-based SSC, as state-of-the-art methods~\cite{liu2025disc, symphonies, li2023voxformer, occformer} initialize 3D geometric priors via back-projecting metric depth maps (see \cref{sec_related_work}).
Notably, no established metric mono depth models exist for the aerial domain.
We therefore evaluate the potential of OccuFly's metric depth maps by benchmarking foundation models in zero-shot and fine-tune settings for MapAnythingV1.1~\cite{keetha2026mapanything}, Metric3Dv2~\cite{hu2024metric3dv2}, and DepthAnything version 2~\cite{yang2024depthAnythingV2} and version~3~\cite{lin2025depthanything3}, using established metrics and official implementations.
Note that MapAnything cannot be fine-tuned, as it is trained for 3D reconstruction and produces depth maps only as a byproduct.
Furthermore, DepthAnything3 does not provide train scripts. We therefore zero-shot evaluate it and fine-tune v2 instead.

\begin{table}[t!]
    \caption{Altitude-wise quantitative SSC evaluation for DISC~\cite{liu2025disc} and Symphonies~\cite{symphonies} on the OccuFly test set, discussed in \cref{subsec_quantitative_results}.}
    \centering
    \resizebox{0.65\linewidth}{!}{
    \begin{tabular}{c|l|c|c}
        \toprule
        {Altitude \SI{}{[m]}} 
            & Method 
            & IoU [\%] 
            & mIoU [\%] \\
        \midrule
        \multirow{2}{*}{{$\mathbf{50}$}}
            & Symphonies~\cite{symphonies} & $15.88$ & $0.58$ \\
            & DISC~\cite{liu2025disc}       & $\mathbf{31.10}$ & $\mathbf{2.20}$ \\
        \midrule
        \multirow{2}{*}{{$\mathbf{40}$}}
            & Symphonies~\cite{symphonies} & $10.71$ & $0.52$ \\
            & DISC~\cite{liu2025disc}       & $\mathbf{27.85}$ & $\mathbf{1.77}$ \\
        \midrule
        \multirow{2}{*}{{$\mathbf{30}$}}
            & Symphonies~\cite{symphonies} & $13.22$ & $0.76$ \\
            & DISC~\cite{liu2025disc}       & $\mathbf{26.88}$ & $\mathbf{2.23}$ \\
        \midrule
        \multirow{2}{*}{{$\mathbf{All}$}}
            & Symphonies~\cite{symphonies} & $13.68$ & $0.58$ \\
            & DISC~\cite{liu2025disc}      & $\mathbf{29.52}$ & $\mathbf{2.04}$ \\
        \bottomrule
    \end{tabular}
    }
    \label{alt_wise_SSC_quantitative_results}
\end{table}

\begingroup
\setlength{\tabcolsep}{3pt}
\begin{table}[t!]
\centering
\caption{Depth estimation evaluation, comparing zero-shot vs. fine-tuned foundation models on the OccuFly test set (see \cref{subsec_quantitative_results}).}
\renewcommand{\arraystretch}{0.8}
\resizebox{\linewidth}{!}{
\begin{tabular}{@{}l|ccc|cccc@{}}
\toprule
Method & $\delta_1\uparrow$ & $\delta_2\uparrow$ & $\delta_3\uparrow$ & AbsRel$\downarrow$ & RMSE$\downarrow$ & MAE$\downarrow$ & SILog$\downarrow$ \\
\midrule
MapAnythingV1.1~\cite{keetha2026mapanything} & $0.000$ & $0.000$ & $0.003$ & $0.799$ & $30.068$ & $29.309$ & $\underline{0.069}$ \\
Metric3Dv2~\cite{hu2024metric3dv2} & $0.073$ & $0.208$ & $0.455$ & $0.471$ & $19.578$ & $18.409$ & $0.156$ \\
DepthAnything2~\cite{yang2024depthAnythingV2} & $0.002$ & $0.015$ & $0.059$ & $0.729$ & $28.382$ & $27.392$ & $0.192$ \\
DepthAnything3~\cite{lin2025depthanything3} & $0.000$ & $0.005$ & $0.141$ & $0.591$ & $22.615$ & $22.019$ & $\mathbf{0.043}$ \\
\midrule
Metric3Dv2-OccuFly & $\underline{0.278}$ & $\underline{0.806}$ & $\underline{0.985}$ & $\underline{0.381}$ & $\underline{13.643}$ & $\underline{13.134}$ & $0.098$ \\
DepthAnything2-OccuFly & $\mathbf{0.834}$ & $\mathbf{0.976}$ & $\mathbf{0.999}$ & $\mathbf{0.134}$ & $\mathbf{4.844}$ & $\mathbf{4.193}$ & $0.112$ \\
\bottomrule
\end{tabular}
}
\label{tab_depth_evaluation_altitudes}
\end{table}
\endgroup

\subsection{Quantitative Results}
\label{subsec_quantitative_results}

\textbf{Aerial Semantic Scene Completion.}\quad Quantitative results in \cref{alt_wise_SSC_quantitative_results} indicate that state-of-the-art models can recover coarse scene geometry, yet struggle to infer semantic structure.
Our detailed altitude and class-wise analysis provided in the supplementary material further substantiates these observations.
Overall, our findings highlight key challenges specific to aerial SSC and position OccuFly as a robust and valuable benchmark for advancing research in this domain.

\noindent\textbf{Metric Monocular Depth Estimation.}\quad
Results in \cref{tab_depth_evaluation_altitudes} show that foundation models underperform significantly in aerial environments, but improve markedly when fine-tuned on OccuFly depth maps.
Notably, the SILog metric, which captures scale-invariant performance (\ie, relative, non-metric accuracy), shows a slight degradation after fine-tuning, indicating a minor trade-off between relative consistency and metric correctness.
Altitude-wise evaluation in the supplementary reveals a steady performance decline with increasing altitude, suggesting that higher viewpoints introduce greater challenges for accurate depth estimation.
Taken together, these findings demonstrate that altitude is a critical factor in aerial depth prediction and that targeted adaptation can substantially improve performance, establishing OccuFly as a significant contribution for advancing metric 3D understanding in aerial vision.

\subsection{Qualitative Results}
\label{subsec_qualtitative_results}

\textbf{Aerial Semantic Scene Completion.}\quad
Qualitative results in \cref{fig_qualitative_results} substantiate the quantitative findings, showing that models capture coarse geometric layout but fail to produce consistent and accurate semantic predictions. 
These shortcomings are particularly evident in complex aerial viewpoints, reflecting a domain gap for models primarily developed on terrestrial data.
By exposing these limitations, OccuFly highlights key challenges in aerial SSC and provides a rigorous testbed to drive progress in this domain.

\noindent\textbf{Metric Monocular Depth Estimation.}\quad
Qualitative depth visualizations in \cref{fig_qualitative_depth} utilize a single, metrically consistent colormap across all depth maps, allowing for a direct comparison of absolute ranges.
Fine-tuned DepthAnything2-OccuFly (DA2-OccuFly) visually reconstitutes the ground-truth topology with coherent depth gradients and realistic color distributions, indicating well-calibrated metric estimates.
In contrast, zero-shot DepthAnything2 (DA2) often shows sharper object boundaries but is dominated by saturated red hues, evidencing a systematic underestimation of distance and poor absolute scaling.
This visual mismatch mirrors the quantitative results, underscoring that viewpoint altitude and in-domain fine-tuning are key to achieving accurate metric aerial depth estimation.

\section{Conclusion}
\label{sec_limitations}

OccuFly introduces the first real-world aerial 3D SSC benchmark, comprising $9$ scenes and over \SI{20}{k} samples with RGB images, semantic voxel grids, and metric depth maps across $21$ semantic classes. Our LiDAR-free, image-based data framework is highly scalable, requiring minimal manual annotation and fostering holistic aerial scene understanding.
 
Our data generation framework faces challenges that offer research opportunities: (1) It assumes static scenes, suppressing truly dynamic objects.
Dynamic-capable reconstruction methods (\eg, 4D Gaussian Splatting~\cite{song2025coda}) are a promising remedy. (2) Data acquisition may incur temporal inconsistencies across cross-altitude capture, when frustum culling uses images taken later.
(3) The labeling process could be fully automated by replacing manually annotated masks with robust 2D pseudo-labels, further scaling data generation.

{
    \small
    \bibliographystyle{ieeenat_fullname}
    \bibliography{main}
}
\clearpage
\setcounter{page}{1}
\maketitlesupplementary

\begin{figure*}[h!]
    \centering
	\includegraphics[width=0.90\textwidth]{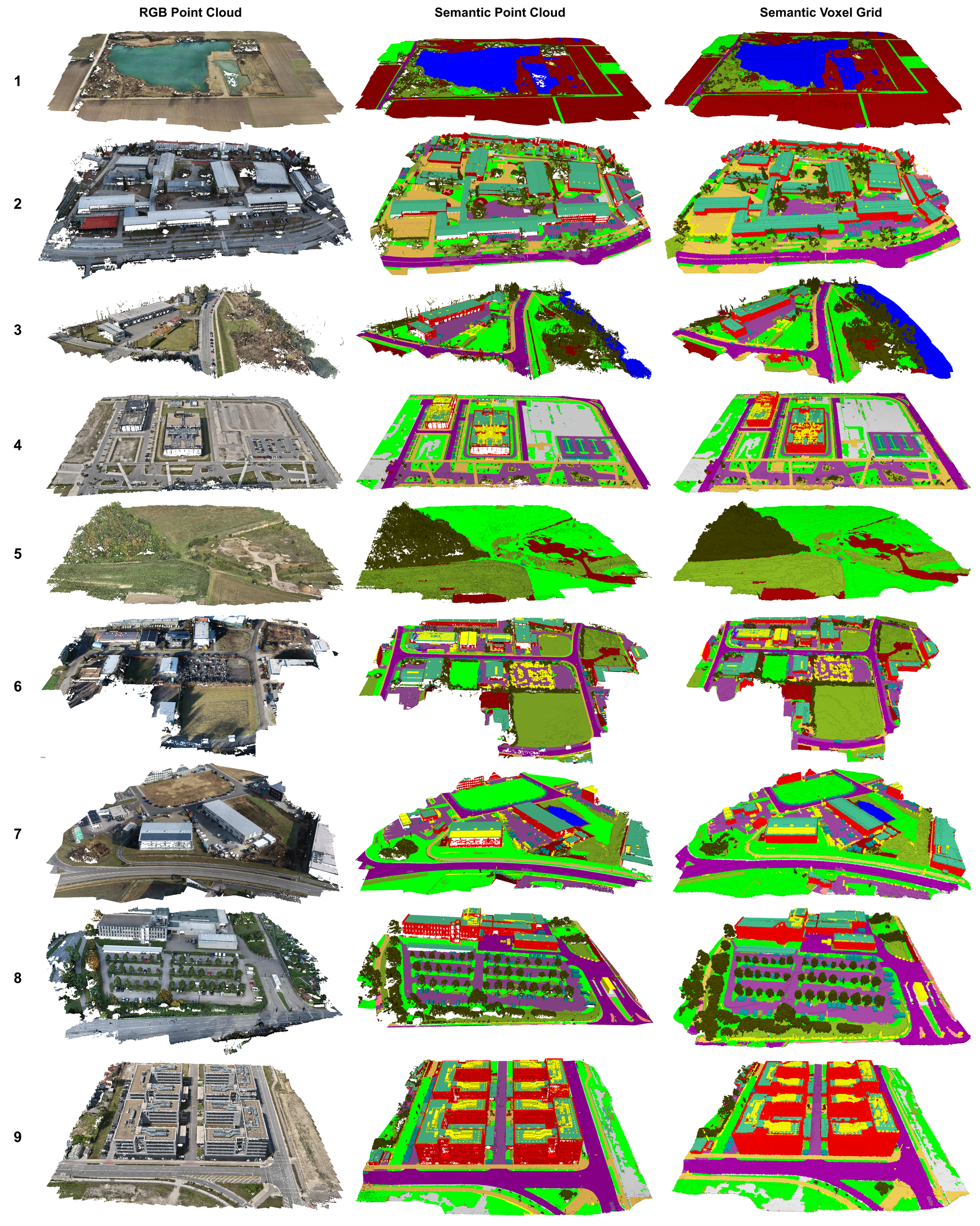}
	\caption{Scene-level outputs of our proposed data generation framework for all scenes 1-9 of the OccuFly dataset. \textbf{Left:} RGB pointcloud from 3D reconstruction (\cref{subsubsec_3d_reconstruction}). \textbf{Center:} Semantic point cloud from semantic annoation (\cref{subsubsec_semantic_annotation}). \textbf{Right:} Semantic voxel grid from densification and voxelization (\cref{subsubsec_voxelization}). Zoom in for best view.}
	\label{supp_fig_rgb_semantic_pcls_and_voxel_grid}    
\end{figure*}

\section{Scene-Level Visualizations}
\label{supp_subsec_implementation_details}
\cref{supp_fig_rgb_semantic_pcls_and_voxel_grid} illustrates (i) the reconstructed RGB point cloud, (ii) the semantic point cloud after label lifting, and (iii) the resulting scene-level semantic voxel grid.
These visualizations demonstrate the remarkable fidelity of our data generation framework and its ability to propagate sparse 2D annotations to a globally consistent, voxel-level 3D ground-truth.

\section{Implementation Details}
\label{supp_subsec_implementation_details}

\paragraph{Data Generation.}
We generate data on an AMD Ryzen Threadripper PRO 7985WX 64-Cores (allocating 8 cores) with \SI{120}{GB} of memory.

For \textbf{3D reconstruction} (\cref{subsubsec_3d_reconstruction}), we utilize the Agisoft Metashape 2.2.0 photogrammetric reconstruction software.
After reconstruction, we ensure high geometric fidelity across the whole scene by removing a small, noisy margin at the border of the scene, which naturally arises from aerial SfM+MVS due to decreased image overlap (shown in \cref{supp_fig_image_overlap}).
Subsequently, during \textbf{semantic annotation} (\cref{subsubsec_semantic_annotation}), we partition the ground plane into a regular grid of scene-dependent square cells with edge lengths of $23$ to $28$ meters to determine the subset of frames for manual annotation. 
Moreover, we apply kNN with $k=100$ for unlabelled point assignment, and $k=200$ for subsequent label refinement.
Furthermore, for \textbf{densification and voxelization}  (\cref{subsubsec_voxelization}) of \textit{instance classes}, DBSCAN~\cite{ester1996_dbscan} clustering is performed with class-wise parameters detailed in \cref{supp_tab_dbscan_params}, and we set $\alpha = 0.05$ for $\alpha$-Shapes~\cite{edelsbrunner1983ashape} and use $K=24$ camera views during silhouette extraction.
For \textit{ground classes}, we set Poisson reconstruction parameters~\cite{poisson} to a depth of $8$ and a scale of $1.2$. Finally, class-wise group assignments, semantic colors, and class frequencies are reported in \cref{supp_tab_semantic_colors}.

\begin{table}[b!]
\centering
\caption{Class-wise DBSCAN~\cite{ester1996_dbscan} parameters for instance separation, discussed in \cref{subsubsec_voxelization}.}
\resizebox{0.55\linewidth}{!}{
\begin{tabular}{l|cc}
\toprule
Class & $\epsilon$ & $\text{MinPts}$ \\
\midrule
Building & $4.0$ & $1000$ \\
Roof & $1.0$ & $1000$ \\
Vehicle & $1.0$ & $500$ \\
Crane & $1.0$ & $500$ \\
Bicycle & $0.4$ & $80$ \\
Person & $0.3$ & $10$ \\
Truck & $1.0$ & $500$ \\
\bottomrule
\end{tabular}
}
\label{supp_tab_dbscan_params}
\end{table}

\paragraph{Aerial Semantic Scene Completion.}
We follow the training protocols of Symphonies~\cite{symphonies} and DISC~\cite{liu2025disc} using their official implementations from GitHub.
We adapt the codebases to OccuFly’s voxel grid resolution of $192 \times 128 \times 128$.
All experiments are conducted on a single NVIDIA A100 80GB GPU with a batch size of $1$.

\paragraph{Metric Monocular Depth Estimation.}
We employ Map-Anything-v1.1~\cite{keetha2026mapanything}, Metric3D-v2-ViT-L~\cite{hu2024metric3dv2}, Depth-Anything-v2-ViT-Small~\cite{yang2024depthAnythingV2}, and Depth-Anything-v3-Nested-Giant-Large~\cite{lin2025depthanything3} (as this is the only metric variant).
For Depth-Anything-v2 specifically, we follow its metric adaptation protocol and fine-tune the affine-invariant model.
Technically, we use official implementations for all methods and train on a single NVIDIA A100 GPU with 80~GB of memory.

\begin{table}[t]
\centering
\caption{Semantic class frequencies, group assignments (\cref{subsubsec_voxelization}), and semantic color table of the OccuFly dataset.}
\resizebox{0.8\linewidth}{!}{
\begin{tabular}{l c l c}
\toprule
Group & Color & Name & Frequency $[\%]$ \\
\midrule
\multirow{8}{*}{Instance}
  & \semcolor{building}      & Building          & \SI{62.1534}{} \\
  & \semcolor{roof}          & Roof              & \SI{2.2018}{} \\
  & \semcolor{vehicle}       & Vehicle           & \SI{0.5683}{} \\
  & \semcolor{cranes}        & Crane             & \SI{0.0059}{} \\
  & \semcolor{bicycle}       & Bicycle           & \SI{0.0035}{} \\
  & \semcolor{person}        & Person            & \SI{0.0001}{} \\
  & \semcolor{truck}         & Truck             & \SI{0.1105}{} \\
\midrule
\multirow{8}{*}{Ground}
  & \semcolor{grass}         & Grass             & \SI{8.5614}{} \\
  & \semcolor{vegetation}    & Vegetation        & \SI{4.3121}{} \\
  & \semcolor{water}         & Water             & \SI{1.7539}{} \\
  & \semcolor{walkway}       & Walkway           & \SI{2.0610}{} \\
  & \semcolor{dirt}          & Dirt              & \SI{2.3364}{} \\
  & \semcolor{road}          & Road              & \SI{1.8099}{} \\
  & \semcolor{gravel}        & Gravel            & \SI{1.4511}{} \\
  & \semcolor{parkinglot}    & Parking Lot       & \SI{2.8415}{} \\
\midrule
\multirow{6}{*}{Others}
  & \semcolor{tree}          & Tree              & \SI{7.5479}{} \\
  & \semcolor{groundobstacles} & Ground Obstacle & \SI{1.9605}{} \\
  & \semcolor{constructions} & Construction      & \SI{0.2741}{} \\
  & \semcolor{cabletower}    & Cable Tower       & \SI{0.0047}{} \\
  & \semcolor{rock}          & Rock              & \SI{0.0402}{} \\
  & \semcolor{cables}        & Cable             & \SI{0.0018}{} \\
\bottomrule
\end{tabular}
}
\label{supp_tab_semantic_colors}
\end{table}

\section{Additional OccuFly Dataset Evaluation}
\label{supp_sec_additional_dataset_evaluation}

\paragraph{3D Reconstruction.}

We provide scene-wise reprojection errors in \cref{supp_tab_reprojection_errors}. An average root mean square reprojection error of \SI{1.24}{pixels} in our geo-referenced images validates the high metric accuracy of the reconstructed point cloud. Scene-wise reconstructed point clouds are shown in \cref{supp_fig_rgb_semantic_pcls_and_voxel_grid}.

\begin{table}[t!]
\centering
\caption{Scene-wise root mean square (RMS) reprojection error after 3D reconstruction (\cref{subsubsec_3d_reconstruction}).}
\resizebox{0.7\linewidth}{!}{
\begin{tabular}{c|c}
\toprule
Scene & RMS Reprojection Error \SI{}{[px]} \\
\midrule
$1$ & $0.469$ \\
$2$ & $0.474$  \\
$3$ & $0.388$  \\
$4$ & $0.451$  \\
$5$ & $0.422$  \\
$6$ & $2.13$   \\
$7$ & $2.04$   \\
$8$ & $2.61$   \\
$9$ & $2.22$  \\
\midrule
Average & $1.24$ \\
\bottomrule
\end{tabular}
}
\label{supp_tab_reprojection_errors}
\end{table}

\begin{figure}[t]
	\centering
	\includegraphics[width=.98\linewidth]{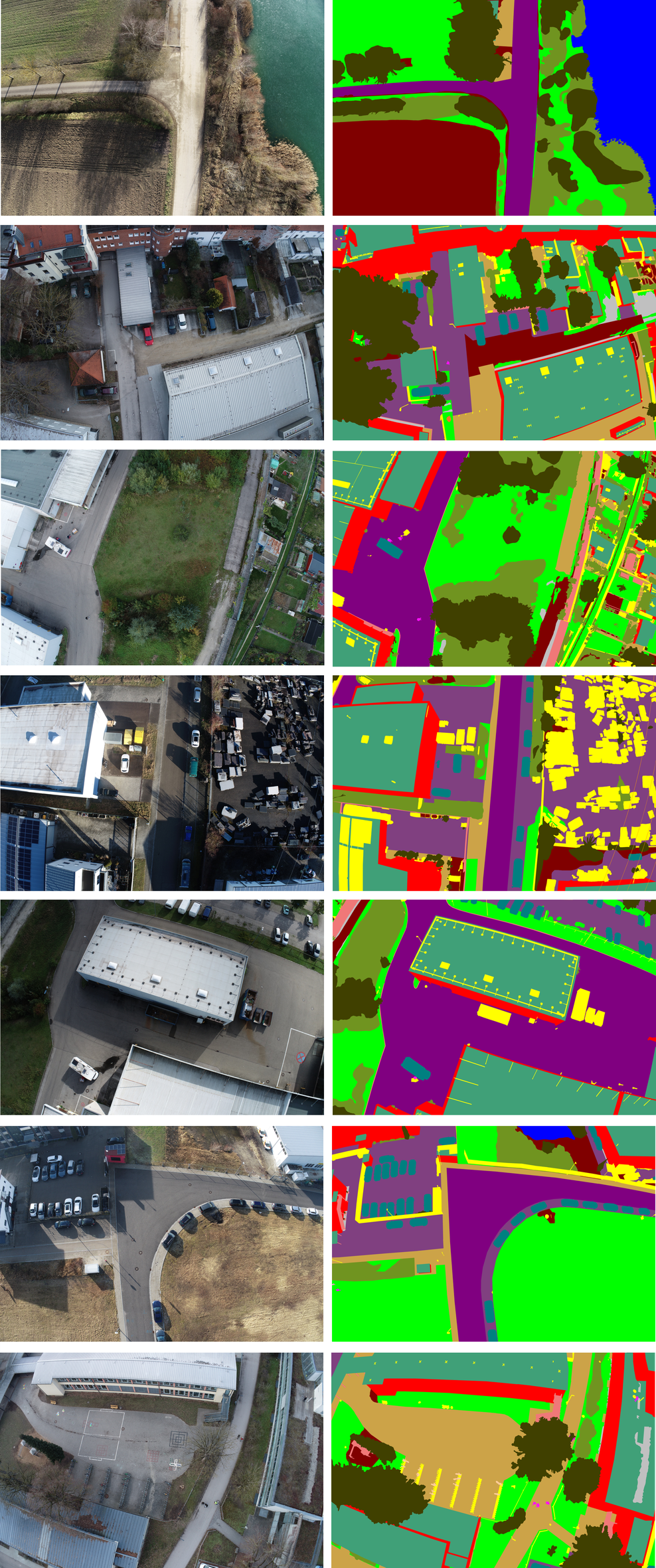}
	\caption{Examples of manual image annotations used for 3D label lifting, showing exceptional pixel-accurate delineation. Zoom in for best view.}
	\label{supp_fig_2d_semantic_maps}    
\end{figure}

~ \\
\paragraph{Semantic Annotation.}
As detailed in \cref{subsubsec_semantic_annotation}, we manually annotate only a small subset of images to subsequently lift semantic labels to 3D. 
This annotation costs approximately 29 minutes per image, which results in 1.3 days per scene to annotate \textgreater \mbox{1.5 billion} voxels in total.
Due to our coverage-aware selection of annotated images, manual effort grows sublinearly with the number of 3D points.
In \cref{supp_fig_2d_semantic_maps}, we present qualitative examples of the manual annotations, which exhibit pixel-accurate delineation.
Furthermore, \cref{supp_tab_annotation_efficiency} reports per-scene annotation ratios, achieving an average annotation ratio of \SI{9.17}{\percent}, indicating exceptional annotation efficiency.
Additionally, \cref{supp_fig_image_overlap} shows scene-wise image overlap during data collection, which exceeds \textgreater\SI{90}{\%} for all scenes. This substantial overlap ensures accurate 3D reconstruction and semantic label lifting. 

\begin{table}[t!]
\centering
\caption{Scene-wise manual semantic annotation ratios for UAV platforms DJI Phantom 4 RTK (P4)~\cite{dji2025phantom4rtk} and DJI Mavic 3 Enterprise Series (M3-ES)~\cite{dji2025mavic3enterprise}. Note that the number of acquired images marginally differs from the number of images finally provided in the dataset, as we remove images at the border of each reconstructed scene to ensure high geometric fidelity (see \cref{supp_subsec_implementation_details}).}
\resizebox{0.8\linewidth}{!}{
\begin{tabular}{c|c|c|c|c}
\toprule
 & UAV & Aquired & Annotated & Ratio \\
Scene & Platform & Images & Images & $[\%]$ \\
\midrule
$1$ & P4 & $421$ & $73$  & $17.34$ \\
$2$ & P4 & $338$ & $48$  & $14.20$ \\
$3$ & M3-ES & $1048$ & $66$ & $6.30$ \\
$4$ & M3-ES & $1252$ & $102$ & $8.15$ \\
$5$ & M3-ES & $1082$ & $74$ & $6.84$ \\
$6$ & P4 & $380$ & $52$  & $13.6$8 \\
$7$ & P4 & $284$ & $40$  & $14.0$8 \\
$8$ & P4 & $251$ & $38$  & $15.14$ \\
$9$ & M3-ES & $1337$ & $93$ & $6.96$ \\
\midrule
\textbf{Total} & & $\mathbf{6393}$ & $\mathbf{586}$ & $\mathbf{9.17}$ \\
\bottomrule
\end{tabular}
}
\label{supp_tab_annotation_efficiency}
\end{table}

\begin{figure}[t!]
	\centering
	\includegraphics[width=.99\linewidth]{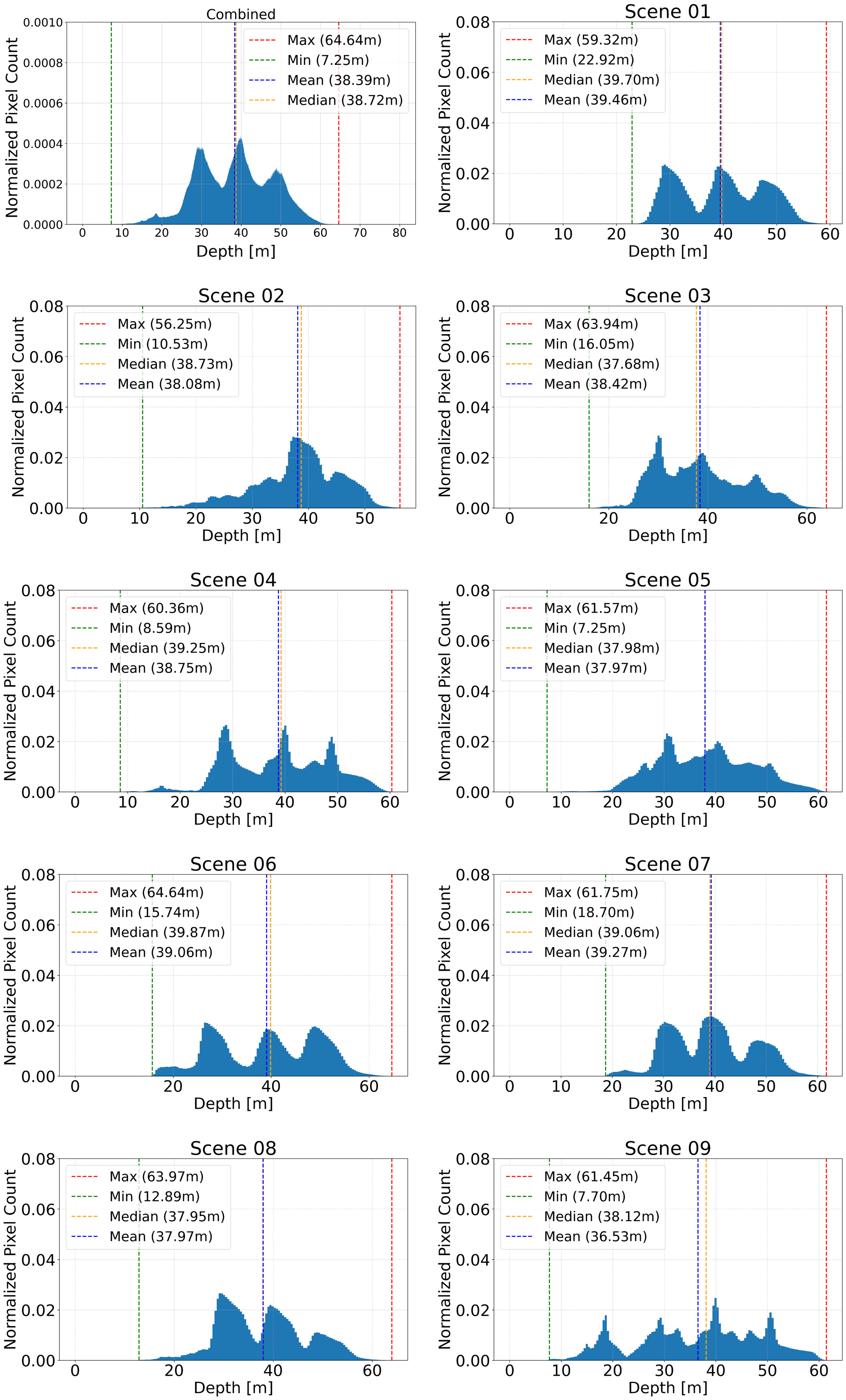}
	\caption{Depth map histograms for each of the 9 scenes in the OccuFly dataset. Zoom in for best view.}
	\label{supp_fig_depth_histograms}    
\end{figure}

\clearpage
\paragraph{Metric Depth Maps.}
\Cref{supp_tab_reprojection_errors} reports a mean reprojection error of \SI{1.24}{pixels}, indicating high geometric consistency of the reconstruction. Since the metric depth maps are derived from these reconstructed points, a low reprojection error serves as a strong proxy for depth accuracy.
Moreover, we provide per-scene depth histograms for all nine scenes to illustrate the dataset’s metric depth distributions (see \cref{supp_fig_depth_histograms}). Most scenes show peaks at 30–50 meters, reflecting the image acquisition altitudes, while certain scenes, such as scene 9, exhibit a more diverse depth range.

\paragraph{Semantic Class Taxonomy.} 
Beyond SSC, \cref{tab_supp_num_semantic_classes} positions OccuFly among established 2D aerial semantic segmentation datasets, where its 21-class taxonomy ranks second. While SkyScapes~\cite{azimi2019skyscapes} ranks first, 12 of its 31 classes are lane-markings, effectively reducing its distinct class count to 20. Consequently, OccuFly provides one of the most detailed aerial taxonomies to date, strengthening fine-grained semantic evaluation and enabling seamless comparability with established 2D benchmarks.

\begin{figure}[t!]
	\centering
	\includegraphics[width=.99\linewidth]{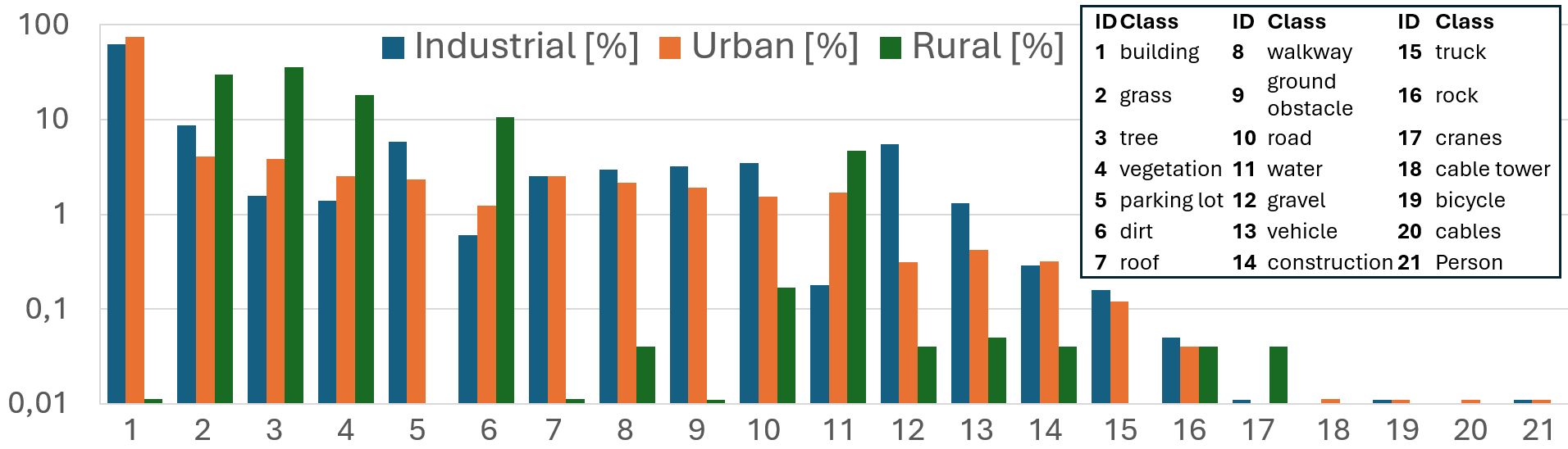}
	\caption{Environment-wise semantic class frequencies.}
	\label{supp_scenario_wise_semantic_class_freq}    
\end{figure}

\paragraph{Dataset Diversity.}
The class distributions shown in \cref{supp_scenario_wise_semantic_class_freq} highlight distinct semantic characteristics across environments, indicating that OccuFly provides rich domain diversity with distinct spatial layouts, architectural densities, and scale, by varying season, environment, altitude, and disjoint geographic locations across splits.

\begin{table}[t!]
\centering
\caption{Semantic class comparison with 2D aerial image datasets.}
\resizebox{\linewidth}{!}{
\begin{tabular}{l|c|c|c|c|c|c|c}
\toprule
& \makecell{UDD\\\cite{chen2018udd}} 
& \makecell{VDD\\\cite{cai2025vdd}} 
& \makecell{UAVid\\\cite{lyu2020uavid}} 
& \makecell{AeroScapes\\\cite{nigam2018aeroscapes}} 
& \makecell{ICG\\\cite{icgDataset}} 
& \makecell{SkyScapes\\\cite{azimi2019skyscapes}} 
& \makecell{OccuFly\\(ours)} \\
\midrule
\# Classes & 4 & 7 & 8 & 11 & 20 & 31 (20) & 21 \\
\bottomrule
\end{tabular}
}
\label{tab_supp_num_semantic_classes}
\end{table}

\section{Additional Benchmark Evaluation}
\label{supp_sec_additional_benchmark_evaluation}

\subsection{Aerial Semantic Scene Completion}

\begin{figure*}[t!]
	\centering
	\includegraphics[width=.99\textwidth]{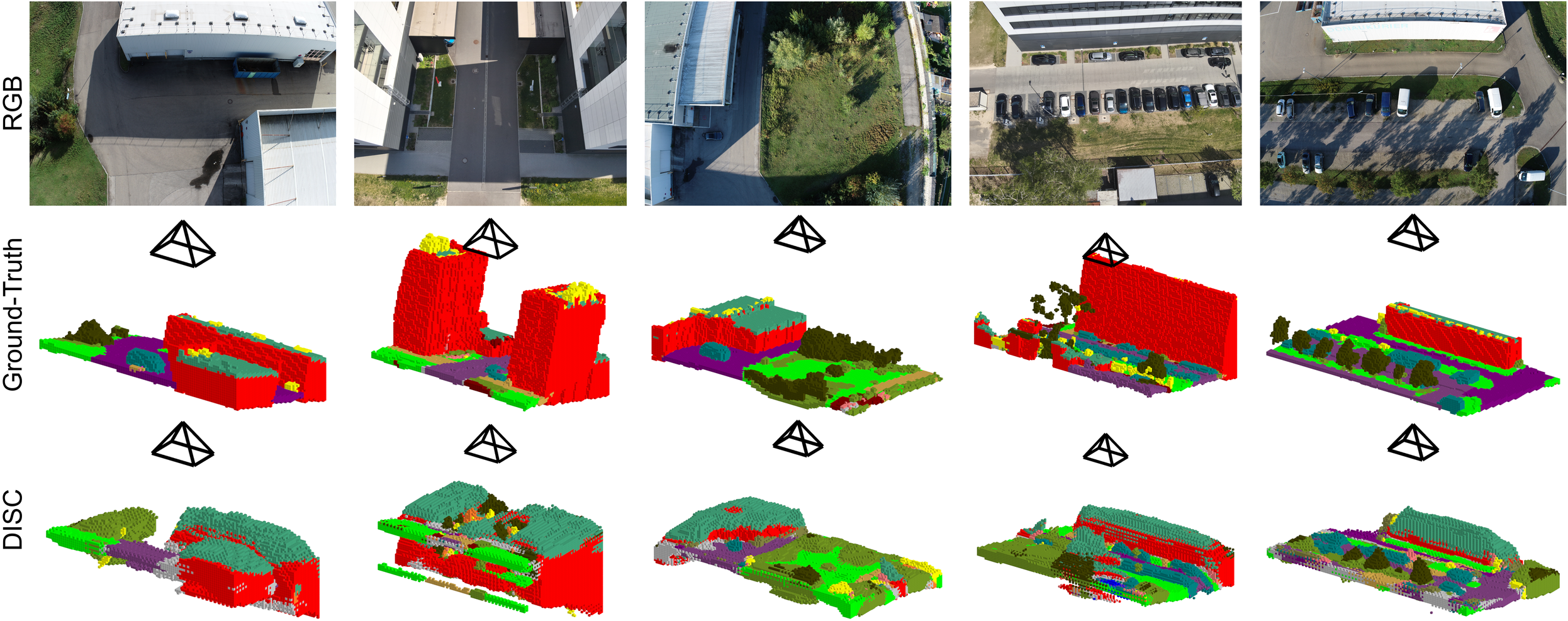}
	\caption{Additional SSC visualizations and qualitative evaluation of DISC~\cite{liu2025disc} on the OccuFly test set.}
	\label{supp_fig_qualitative_results}    
\end{figure*}

In addition to the evaluation discussed in \cref{subsec_quantitative_results,subsec_qualtitative_results}, we provide altitude-wise and class-wise metrics in \cref{supp_fig_class_wise_quantitative_results}, along with further qualitative results in \cref{supp_fig_qualitative_results}. Consistent with the main manuscript, performance remains uniformly low across all altitude ranges, indicating that viewpoint height has a limited impact compared to the overall difficulty of the task. The class-wise analysis further reflects this trend, with only frequent classes being recovered to a limited extent, while rare classes are largely undetected. Qualitative results underscore these observations, showing coarse geometric structure but fragmented semantic predictions. Together, these findings reinforce the challenges of aerial SSC and underline the need for dedicated benchmarks such as OccuFly.

\begin{table*}[t]
    \caption{Altitude and class-wise SSC performance for Symphonies~\cite{symphonies} and DISC~\cite{liu2025disc} on the OccuFly test set in \% (\textbf{best}).}
    \begin{subtable}{1.0\textwidth}
        \scriptsize
        \setlength{\tabcolsep}{0.004\linewidth}
        \centering
        \resizebox{0.99\textwidth}{!}{%
        \begin{tabular}{c|l|c c|c c c c c c c c c c c c c c c c c c c c  c}
            \toprule
            \rotatebox{90}{Altitude \SI{}{[m]}}
            & Method
            & \rotatebox{90}{IoU}
            & \rotatebox{90}{mIoU}
            & \rotatebox{90}{\textcolor{road}{$\blacksquare$} Road\classfreq{road}}
            & \rotatebox{90}{\textcolor{walkway}{$\blacksquare$} Walkway\classfreq{walkway}}
            & \rotatebox{90}{\textcolor{dirt}{$\blacksquare$} Dirt\classfreq{dirt}}
            & \rotatebox{90}{\textcolor{gravel}{$\blacksquare$} Gravel\classfreq{gravel}}
            & \rotatebox{90}{\textcolor{rock}{$\blacksquare$} Rock\classfreq{rock}}
            & \rotatebox{90}{\textcolor{grass}{$\blacksquare$} Grass\classfreq{grass}}
            & \rotatebox{90}{\textcolor{vegetation}{$\blacksquare$} Vegetation\classfreq{vegetation}}
            & \rotatebox{90}{\textcolor{tree}{$\blacksquare$} Tree\classfreq{tree}}
            & \rotatebox{90}{\textcolor{groundobstacles}{$\blacksquare$} Ground-Obs.\classfreq{groundobstacles}}
            & \rotatebox{90}{\textcolor{person}{$\blacksquare$} Person\classfreq{person}}
            & \rotatebox{90}{\textcolor{bicycle}{$\blacksquare$} Bicycle\classfreq{bicycle}}
            & \rotatebox{90}{\textcolor{vehicle}{$\blacksquare$} Vehicle\classfreq{vehicle}}
            & \rotatebox{90}{\textcolor{water}{$\blacksquare$} Water\classfreq{water}}
            & \rotatebox{90}{\textcolor{building}{$\blacksquare$} Building\classfreq{building}}
            & \rotatebox{90}{\textcolor{roof}{$\blacksquare$} Roof\classfreq{roof}}
            & \rotatebox{90}{\textcolor{cables}{$\blacksquare$} Cables\classfreq{cables}}
            & \rotatebox{90}{\textcolor{cabletower}{$\blacksquare$} Cable-Tower\classfreq{cabletower}}
            & \rotatebox{90}{\textcolor{parkinglot}{$\blacksquare$} Parking-Lot\classfreq{parkinglot}}
            & \rotatebox{90}{\textcolor{constructions}{$\blacksquare$} Constructions\classfreq{constructions}}
            & \rotatebox{90}{\textcolor{cranes}{$\blacksquare$} Cranes\classfreq{cranes}}
            & \rotatebox{90}{\textcolor{truck}{$\blacksquare$} Truck\classfreq{truck}}
            \\
            \midrule
            \multirow{2}{*}{\rotatebox{90}{${50}$}} 
            & Symphonies
            & $15.88$ & $0.58$ & $1.22$ & $0.15$ & $\mathbf{0.19}$
            & $\mathbf{0.54}$ & $0.01$ & $1.64$ & $0.61$
            & $\mathbf{0.72}$ & $0.20$ & $0.00$ & $0.00$
            & $0.35$ & $0.00$ & $4.66$ & $0.75$
            & $0.00$ & $0.00$ & $0.00$ & $0.00$ & $0.00$ & $0.00$ \\
            & DISC
            & $\mathbf{31.10}$ & $\mathbf{2.20}$ & $\mathbf{1.24}$ & $\mathbf{0.74}$ & $0.12$
            & $0.24$ & $0.01$ & $\mathbf{2.62}$ & $\mathbf{0.80}$
            & $0.23$ & $\mathbf{1.37}$ & $0.00$ & $\mathbf{0.01}$
            & $\mathbf{4.37}$ & $0.00$ & $\mathbf{28.58}$ & $\mathbf{1.38}$
            & $0.00$ & $0.00$ & $0.00$ & $0.00$ & $0.00$ & $0.00$ \\
            \midrule
            \multirow{2}{*}{\rotatebox{90}{${40}$}} 
            & Symphonies
            & $10.71$ & $0.52$ & $0.40$ & $0.09$ & $0.05$
            & $0.02$ & $0.01$ & $1.59$ & $1.08$
            & $\mathbf{1.92}$ & $0.09$ & $0.00$ & $0.00$
            & $0.68$ & $0.00$ & $3.38$ & $0.57$
            & $0.00$ & $0.00$ & $0.00$ & $0.00$ & $0.00$ & $0.00$ \\
            & DISC
            & $\mathbf{27.85}$ & $\mathbf{1.77}$ & $\mathbf{0.87}$ & $\mathbf{0.25}$ & $\mathbf{0.18}$
            & $\mathbf{0.13}$ & $\mathbf{0.02}$ & $\mathbf{1.61}$ & $\mathbf{1.54}$
            & $1.42$ & $\mathbf{0.28}$ & $0.00$ & $0.00$
            & $\mathbf{4.80}$ & $0.00$ & $\mathbf{21.46}$ & $\mathbf{1.12}$
            & $0.00$ & $0.00$ & $0.00$ & $0.00$ & $0.00$ & $0.00$ \\
            \midrule
            \multirow{2}{*}{\rotatebox{90}{${30}$}} 
            & Symphonies
            & $13.22$ & $0.76$ & $0.56$ & $0.41$ & $0.09$
            & $\mathbf{0.25}$ & $0.00$ & $2.82$ & $1.72$
            & $\mathbf{2.32}$ & $0.13$ & $0.00$ & $0.00$
            & $1.19$ & $0.00$ & $3.39$ & $1.60$
            & $0.00$ & $0.00$ & $0.00$ & $0.00$ & $0.00$ & $0.00$ \\
            & DISC
            & $\mathbf{26.88}$ & $\mathbf{2.23}$ & $\mathbf{0.98}$ & $\mathbf{0.44}$ & $\mathbf{0.60}$
            & $0.10$ & $\mathbf{0.01}$ & $\mathbf{3.17}$ & $\mathbf{2.41}$
            & $0.78$ & $\mathbf{0.32}$ & $0.00$ & $0.00$
            & $\mathbf{5.45}$ & $0.00$ & $\mathbf{26.31}$ & $\mathbf{1.78}$
            & $0.00$ & $0.00$ & $0.00$ & $0.00$ & $0.00$ & $0.00$ \\
            \midrule
            \multirow{2}{*}{\rotatebox{90}{all}} 
            & Symphonies
            & $13.68$ & $0.58$ & $0.88$ & $0.14$ & $0.13$
            & $\mathbf{0.33}$ & $0.01$ & $1.76$ & $0.94$
            & $\mathbf{1.42}$ & $0.15$ & $0.00$ & $0.00$
            & $0.51$ & $0.00$ & $4.08$ & $0.74$
            & $0.00$ & $0.00$ & $0.00$ & $0.00$ & $0.00$ & $0.00$ \\
            & DISC
            & $\mathbf{29.52}$ & $\mathbf{2.04}$ & $\mathbf{1.08}$ & $\mathbf{0.53}$ & $\mathbf{0.17}$
            & $0.18$ & $0.01$ & $\mathbf{2.32}$ & $\mathbf{1.33}$
            & $0.77$ & $\mathbf{0.90}$ & $0.00$ & $\mathbf{0.01}$
            & $\mathbf{4.67}$ & $0.00$ & $\mathbf{25.48}$ & $\mathbf{1.30}$
            & $0.00$ & $0.00$ & $0.00$ & $0.00$ & $0.00$ & $0.00$ \\
            \bottomrule
        \end{tabular}
        }
    \end{subtable}
    \label{supp_fig_class_wise_quantitative_results}
\end{table*}

\subsection{Aerial Metric Monocular Depth Estimation}
\label{supp_metric_monocular_depth_estimation}

In addition to Tab.~\ref{tab_depth_evaluation_altitudes}, we provide a detailed analysis of depth estimation performance across different altitude ranges, presented in Tab.~\ref{supp_tab_depth_evaluation_altitudes}.
The observed trends are consistent with the findings in the main manuscript: performance degrades with increasing altitude, highlighting the difficulty of metric depth estimation from higher viewpoints, while fine-tuning leads to substantial improvements across all altitudes.

\begingroup
\setlength{\tabcolsep}{6pt}
\begin{table*}[t!]
\centering
\caption{Altitude-wise metric monocular depth estimation performance, comparing zero-shot vs. fine-tuned foundation models on the OccuFly test set.}
\renewcommand{\arraystretch}{1.0}
\resizebox{0.9\linewidth}{!}{
\begin{tabular}{@{}c|l|ccc|cccc@{}}
\toprule
Altitude \SI{}{[m]} & Method & $\delta_1\uparrow$ & $\delta_2\uparrow$ & $\delta_3\uparrow$ & AbsRel$\downarrow$ & RMSE$\downarrow$ & MAE$\downarrow$ & SILog$\downarrow$ \\
\midrule

\multirow{6}{*}{50} 
& MapAnythingV1.1~\cite{keetha2026mapanything} & $0.000$ & $0.000$ & $0.000$ & $0.836$ & $36.783$ & $35.935$ & $\underline{0.057}$ \\
& Metric3Dv2~\cite{hu2024metric3dv2} & $0.014$ & $0.161$ & $0.363$ & $0.516$ & $24.582$ & $23.250$ & $0.166$ \\
& DepthAnything2~\cite{yang2024depthAnythingV2} & $0.000$ & $0.000$ & $0.003$ & $0.767$ & $34.576$ & $33.515$ & $0.191$ \\
& DepthAnything3~\cite{lin2025depthanything3} & $0.000$ & $0.000$ & $0.000$ & $0.673$ & $29.703$ & $29.017$ & $\mathbf{0.030}$ \\
\cline{2-9}
& Metric3Dv2-OccuFly & $\underline{0.283}$ & $\underline{0.831}$ & $\underline{0.987}$ & $\underline{0.379}$ & $\underline{15.373}$ & $\underline{14.950}$ & $0.099$ \\
& DepthAnything2-OccuFly & $\mathbf{0.844}$ & $\mathbf{0.997}$ & $\mathbf{1.000}$ & $\mathbf{0.129}$ & $\mathbf{5.985}$ & $\mathbf{5.261}$ & $0.114$ \\

\midrule

\multirow{6}{*}{40} 
& MapAnythingV1.1~\cite{keetha2026mapanything} & $0.000$ & $0.001$ & $0.005$ & $0.780$ & $26.735$ & $25.906$ & $\underline{0.083}$ \\
& Metric3Dv2~\cite{hu2024metric3dv2} & $0.137$ & $0.277$ & $0.485$ & $0.437$ & $17.102$ & $15.808$ & $0.165$ \\
& DepthAnything2~\cite{yang2024depthAnythingV2} & $0.002$ & $0.024$ & $0.111$ & $0.693$ & $25.031$ & $23.918$ & $0.208$ \\
& DepthAnything3~\cite{lin2025depthanything3} & $0.000$ & $0.001$ & $0.073$ & $0.564$ & $19.450$ & $18.817$ & $\mathbf{0.051}$ \\
\cline{2-9}
& Metric3Dv2-OccuFly & $\underline{0.209}$ & $\underline{0.802}$ & $\underline{0.981}$ & $\underline{0.395}$ & $\underline{13.138}$ & $\underline{12.625}$ & $0.096$ \\
& DepthAnything2-OccuFly & $\mathbf{0.795}$ & $\mathbf{0.957}$ & $\mathbf{0.998}$ & $\mathbf{0.148}$ & $\mathbf{4.486}$ & $\mathbf{3.823}$ & $0.119$ \\

\midrule

\multirow{6}{*}{30} 
& MapAnythingV1.1~\cite{keetha2026mapanything} & $0.000$ & $0.000$ & $0.002$ & $0.764$ & $23.307$ & $22.952$ & $\underline{0.060}$ \\
& Metric3Dv2~\cite{hu2024metric3dv2} & $0.036$ & $0.129$ & $0.589$ & $0.459$ & $14.518$ & $14.084$ & $0.107$ \\
& DepthAnything2~\cite{yang2024depthAnythingV2} & $0.005$ & $0.029$ & $0.050$ & $0.737$ & $22.961$ & $22.493$ & $0.147$ \\
& DepthAnything3~\cite{lin2025depthanything3} & $0.002$ & $0.026$ & $0.669$ & $0.471$ & $14.494$ & $14.216$ & $\mathbf{0.052}$ \\
\cline{2-9}
& Metric3Dv2-OccuFly & $\underline{0.460}$ & $\underline{0.757}$ & $\underline{0.992}$ & $\underline{0.347}$ & $\underline{10.917}$ & $\underline{10.215}$ & $0.101$ \\
& DepthAnything2-OccuFly & $\mathbf{0.919}$ & $\mathbf{0.982}$ & $\mathbf{0.998}$ & $\mathbf{0.103}$ & $\mathbf{3.108}$ & $\mathbf{2.666}$ & $0.086$ \\

\midrule

\multirow{6}{*}{All} 
& MapAnythingV1.1~\cite{keetha2026mapanything} & $0.000$ & $0.000$ & $0.003$ & $0.799$ & $30.068$ & $29.309$ & $\underline{0.069}$ \\
& Metric3Dv2~\cite{hu2024metric3dv2} & $0.073$ & $0.208$ & $0.455$ & $0.471$ & $19.578$ & $18.409$ & $0.156$ \\
& DepthAnything2~\cite{yang2024depthAnythingV2} & $0.002$ & $0.015$ & $0.059$ & $0.729$ & $28.382$ & $27.392$ & $0.192$ \\
& DepthAnything3~\cite{lin2025depthanything3} & $0.000$ & $0.005$ & $0.141$ & $0.591$ & $22.615$ & $22.019$ & $\mathbf{0.043}$ \\
\cline{2-9}
& Metric3Dv2-OccuFly & $\underline{0.278}$ & $\underline{0.806}$ & $\underline{0.985}$ & $\underline{0.381}$ & $\underline{13.643}$ & $\underline{13.134}$ & $0.098$ \\
& DepthAnything2-OccuFly & $\mathbf{0.834}$ & $\mathbf{0.976}$ & $\mathbf{0.999}$ & $\mathbf{0.134}$ & $\mathbf{4.844}$ & $\mathbf{4.193}$ & $0.112$ \\

\bottomrule
\end{tabular}
}
\label{supp_tab_depth_evaluation_altitudes}
\end{table*}

\begin{figure*}[t!]
	\centering
	\includegraphics[width=.99\textwidth]{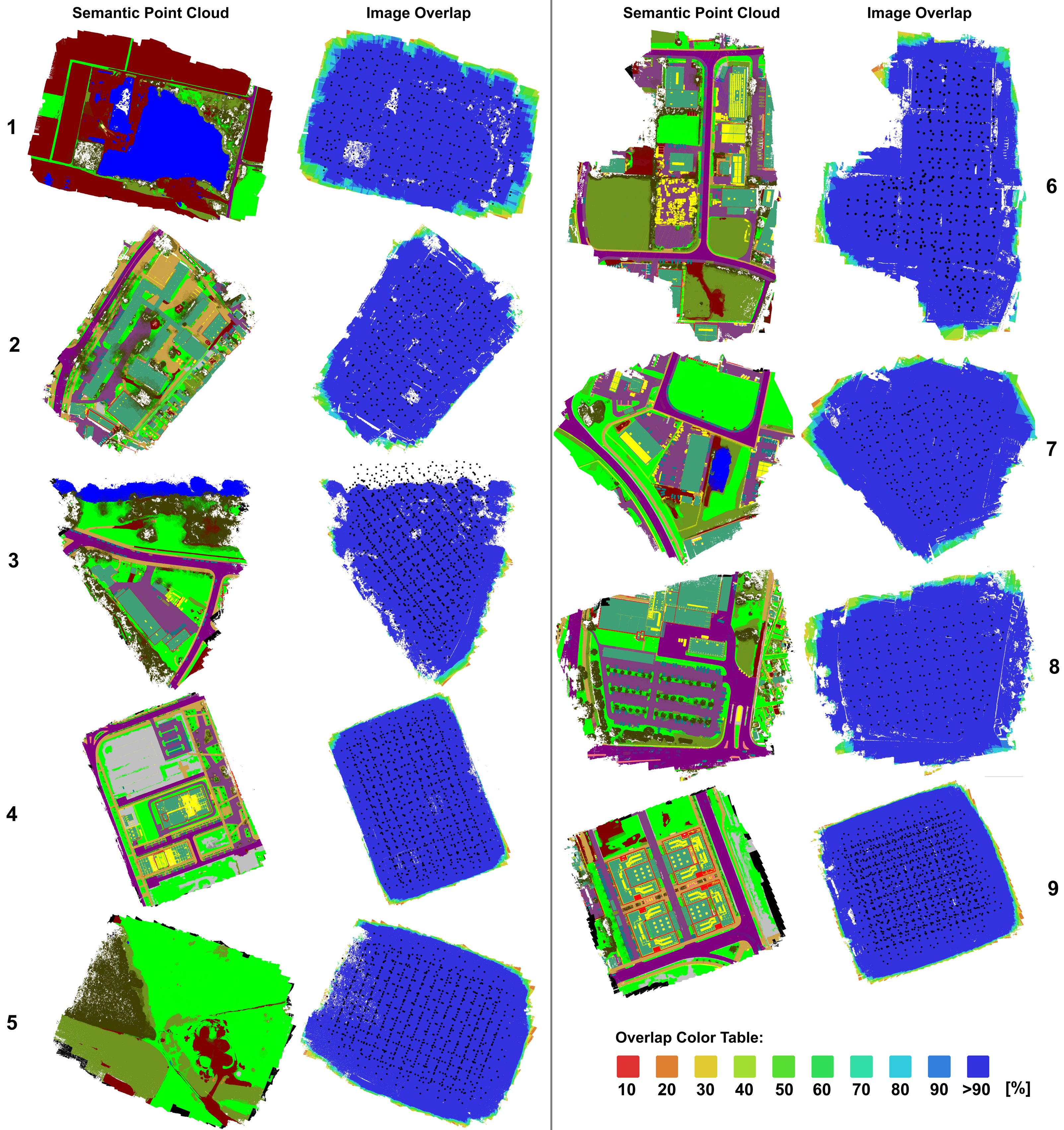}
	\caption{Scene-wise image overlap during data collection for all scenes 1-9 of the OccuFly dataset. \textbf{Left:} Top-down view of the semantic point cloud. \textbf{Right:} Image overlap with camera centers depicted as black dots. Note that we remove scene borders with \SI{<90}{\%} overlap to ensure geometric and semantic fidelity, as discussed in \cref{supp_subsec_implementation_details}. Zoom in for best view.}
	\label{supp_fig_image_overlap}    
\end{figure*}

\end{document}